\pdfoutput=1

\documentclass[11pt]{article}

    \makeatletter
\def\@fnsymbol#1{\ensuremath{\ifcase#1\or \dagger\or \ddagger\or
   \mathsection\or \mathparagraph\or \|\or **\or \dagger\dagger
   \or \ddagger\ddagger \else\@ctrerr\fi}}
    \makeatother
    
\usepackage[table]{xcolor}
\definecolor{maroon}{cmyk}{0,0.87,0.68,0.32}
\usepackage{EMNLP2022}

\usepackage{times}
\usepackage{latexsym}

\usepackage[T1]{fontenc}

\usepackage[utf8]{inputenc}

\usepackage{microtype}

\usepackage{inconsolata}
\usepackage{threeparttable}
\usepackage{multirow}
\usepackage{graphicx}
\usepackage{amsthm}
\usepackage{amssymb}
\usepackage{mathtools}
\usepackage{amsfonts}       
\usepackage{bm}
\usepackage{float}
\usepackage{makecell}
\usepackage{booktabs}
\usepackage{multicol}
\RequirePackage{algorithm}
\RequirePackage{algorithmic}
\newtheorem{proposition}{Proposition}
\newtheorem{definition}{Definition}
\newtheorem{theorem}{Theorem}
\newtheorem{corollary}{Corollary}[theorem]

\newtheorem*{theorem_}{Theorem}
\newtheorem*{corollary_}{Corollary}

\newtheorem*{remark}{Remark}
\usepackage[draft]{changes}
\definechangesauthor[name={Clara}, color=orange]{clara}

\def\eqref#1{equation~\ref{#1}}

\def\1{\bm{1}}

\def\rvc{{\mathbf{c}}}

\def\rvh{{\mathbf{h}}}

\def\rvx{{\mathbf{x}}}
\def\rvy{{\mathbf{y}}}
\def\rvz{{\mathbf{z}}}

\def\rmH{{\mathbf{H}}}

\def\rmN{{\mathbf{N}}}

\def\sQ{{\mathbb{Q}}}

\def\sS{{\mathbb{S}}}

\title{Reconciliation of Pre-trained Models and Prototypical Neural Networks
\\in Few-shot Named Entity Recognition}

\author{
Youcheng Huang$^{\spadesuit\heartsuit}$, \quad 
Wenqiang Lei$^{\spadesuit}$\thanks{ \quad Correspondence to Wenqiang Lei.}, 
\quad Jie Fu$^{\clubsuit}$, 
\quad Jiancheng Lv$^{\spadesuit}$ \\
        ${\spadesuit}$ College of Computer Science, Sichuan University \\
        ${\clubsuit}$ Beijing Academy of Artificial Intelligence \\
        ${\heartsuit}$ laerster@gmail.com \quad $\dagger$ wenqianglei@gmail.com
}
        
\begin{document}
\maketitle

\begin{abstract}
Incorporating large-scale pre-trained models with the prototypical neural networks is a \textit{de-facto} paradigm in few-shot named entity recognition.
Existing methods, unfortunately, are not aware of the fact that embeddings from pre-trained models contain a prominently large amount of information regarding word \textbf{\textit{frequencies}}, biasing prototypical neural networks against learning word \textbf{\textit{entities}}.
This discrepancy constrains the two models' synergy.
Thus, we propose a one-line-code normalization method to reconcile such a mismatch with empirical and theoretical grounds.
Our experiments based on nine benchmark datasets show the superiority of our method over the counterpart models and are comparable to the state-of-the-art methods.
In addition to the model enhancement, our work also provides an analytical viewpoint for addressing the general problems in few-shot name entity recognition or other tasks that rely on pre-trained models or prototypical neural networks.\footnote{Our code is available at \url{https://github.com/HamLaertes/EMNLP_2022_Reconciliation}}
\end{abstract}

\section{Introduction}

Named entity recognition (NER) is a classical task in natural language processing (NLP) which aims to automatically identify entities in the plain text by classifying each word to a set of pre-defined entities, \emph{e.g.} ``person/location'', or to the ``others'' (no-entity) \cite{yadav2019survey}.
As a crucial sub-component of many language understanding tasks, NER has been widely adopted to different applications, \emph{e.g.} news \cite{sang2003introduction} and the medical \cite{stubbs2015annotating}.

Neural networks (NNs) have achieved great success in NER \cite{lample-etal-2016-neural}.
However, NNs face the adaptation challenge \cite{wilson2020survey} as words in different entities can change to a great extent \cite{yang-katiyar-2020-simple}, \emph{e.g.} "Mr. Bush" in the "person" \emph{v.s.} "budgets" in the "money", and obtaining sufficient annotations of new entities can be expensive \cite{ding-etal-2021-nerd}.
Few-shot NER, a cost-efficient solution, aims at training a model to be aware of unseen entities given few labeled examples \cite{huang-etal-2021-shot}.
Few-shot NER has received a rising interest in the NLP community, where new datasets \cite{ding-etal-2021-nerd} and methods \cite{das-etal-2022-container, yang-katiyar-2020-simple, tong-etal-2021-learning} have been constantly proposed.

Low-dim manifold encodes more adaptive information \cite{wang2018visual}.
Prototypical neural networks (PNNs) \cite{snell2017prototypical} learn an embedding space where the same-entity datapoints are clustered around a center, called the prototype, and distances between the query data to all prototypes represent its entity probabilities.
In addition to using an embedding network, PNNs calculate the prototypes and distances {\it via} a {\it non-parameteric} algorithm, gaining popularity for the flexibility and low computing cost \cite{wang2020generalizing}.
A supplementary enhancement will be using embeddings from large-scale pre-trained models (PTMs), like BERT \cite{devlin-etal-2019-bert}, to provide extra knowledge that helps PNNs' learning of entities.
As such, incorporating PTMs with PNNs has become a de-facto paradigm for few-shot NER that achieves competitive results to state-of-the-arts \cite{ding-etal-2021-nerd, huang-etal-2021-shot, bao2020fewshot}.
Related works consider NER-specific properties \cite{tong-etal-2021-learning} or new learning algorithms \cite{das-etal-2022-container, yang-katiyar-2020-simple} to enhance the model, but they tend not to examine the coordinating effects between PTMs and PNNs in terms of the information contained in embeddings.

It should be reminded that PNNs calculate distances between word embeddings and prototypes to represent entity probabilities.
However, PTMs embeddings may not effectively provide entity information as they prominently contain information on word frequencies \cite{mu2018all, li-etal-2020-sentence}, and we find frequencies are shallow statistics that can cause a loss of in-depth and useful entity-denoting information.
By probing into PNNs, we find that words tend to be classified to the entity centred with words of higher frequencies.
Therefore, the distance measure is biased towards focusing on frequencies.
Such a bias can cause the over-fitting of the PNNs and the unreliability on classifying new entities.
As a consequence, when frequencies are changed on a new corpus, the distances can no longer effectively represent the entity probabilities.

Form a mathematical view, the biased distance is mainly caused by the varying prototype $\ell$2-norms.
However, we argue that those $\ell2$-norms contribute little to but actually undermine the correct classification.
We propose to normalize all prototypes to unit vectors as a simple yet effective remedy to reconcile PNNs and PTMs for few-shot NER.
Our experiments on nine few-shot NER datasets \cite{huang-etal-2021-shot, ding-etal-2021-nerd} demonstrate the effectiveness of our one-line-code remedy.
The normalized PNNs achieve competitive results compared to the state-of-the-art methods while retaining all the PNNs' advantages, such as easy implementation and low computation cost.
We also demonstrate normalization can make PNNs learn more effectively about correctly classifying entities, and conduct ablation studies on different normalization strategies.

Our study on reconciling PTMs and PNNs, and the promising performance of the simple normalization method may inspire new research motivations to the few-shot NER, as well as other fields that involve the use of PTMs /or PNNs.

\section{Background and Related Works}

\subsection{Few-shot Classification and Embedding-based Classifiers}
\begin{figure}[!htbp]
    \centering
    \includegraphics[width=0.48\textwidth]{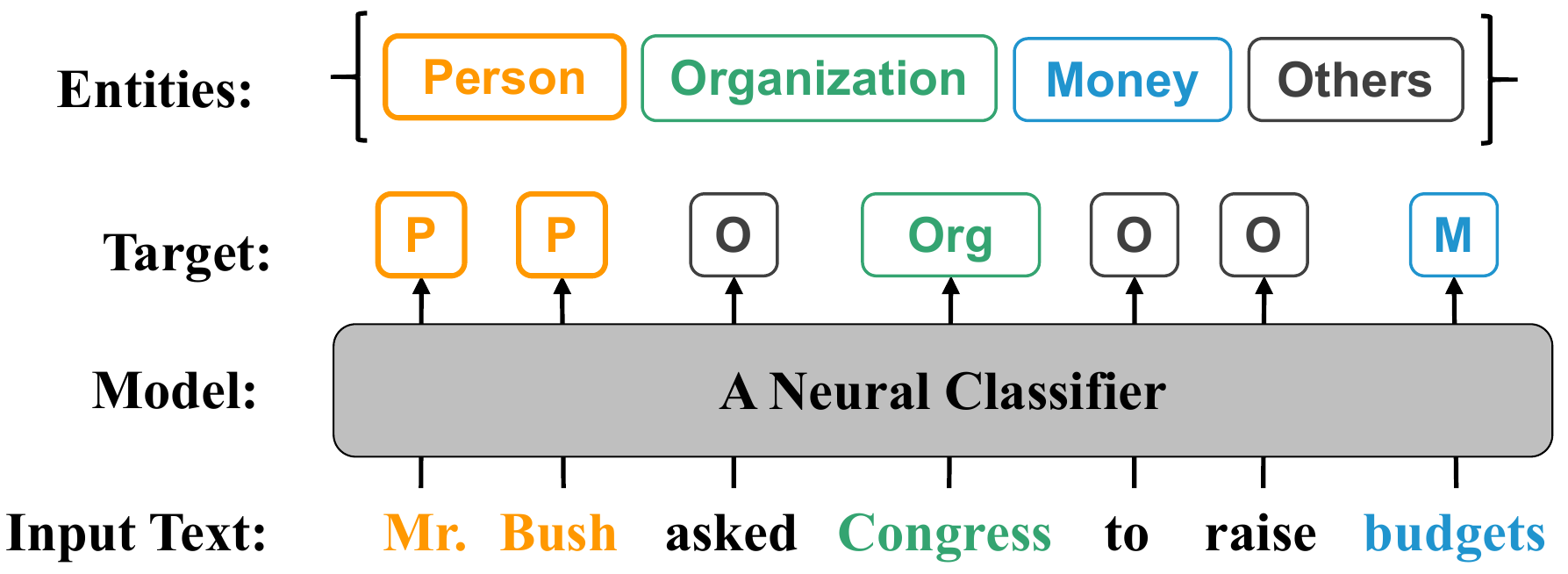}
    \caption{An example of the input and output of NER.}
    \label{fig:nerexample}
\end{figure}
\noindent Named entity recognition can be formalized as the word classification (Figure~\ref{fig:nerexample}).
For few-shot classification (FSC), "$K$-way $N$-shot" describes the task setting: 
after the training, the model needs to classify the query data to $K$ training-unseen classes, given $N$ labeled examples per class.
The core issue in FSC is the unreliable empirical loss minimization:
as the labeled data is extremely limited during testing, the loss defined on new classes will result in a sub-optimal solution that may lead to undesired performance \cite{wang2020generalizing}.

To tackle this issue, researchers seek solutions with the embedding-based methods \cite{wang2020generalizing, koch2015siamese, vinyals2016matching, sung2018learning, snell2017prototypical}.
Specifically, an embedding network projects datapoints to a low-dim manifold that contains some general features shared among training and testing classes.
On the embedding space, to train only a small classifier for new classes consumes fewer data and can achieve equivalently good results.
The recent embedding-based classifiers with meta-learning \cite{hochreiter2001learning} divides the training data into several "episodes" mimicking the "$K$-way $N$-shot" testing format.
Such a method is popularly known for its effectiveness in FSC.

\subsection{Prototypical Neural Network}

PNNs \cite{snell2017prototypical} assume in the embedding space, the same-class datapoints are clustered around class centers, called the prototypes, and the distances between datapoints to prototypes represent their class probabilities.
Based on this assumption, PNNs need only calculate: 1) the prototypes using the embedded labeled data, and 2) the distances between the embedded query data and prototypes to conduct the classification.
The detailed discussions about PNNs will be presented in section~\ref{sec:pnn_assumption} and~\ref{sec:pnn_degeneration}.
Utilizing large-scale PTMs as the embedding networks, PNNs can achieve competitive results in various natural language FSC tasks \cite{ding-etal-2021-nerd, holla2020learning, huang-etal-2021-shot, bao2020fewshot}.

In NER, recent works consider a bunch of methods to enhance the coordinating usage of PTMs and PNNs, including in-domain pre-training \cite{huang-etal-2021-shot}, NER specific properties \cite{tong-etal-2021-learning}, and sophisticated learning algorithm \cite{das-etal-2022-container, yang-katiyar-2020-simple}.
However, to best of our knowledge, little has been explored for the correct combination of PTMs and PNNs.
There have been works that find both the small-scale \cite{Mikolov2013EfficientEO, pennington-etal-2014-glove} and recent large-scale \cite{devlin-etal-2019-bert, liu2019roberta} PTMs have limitations in representing diverse language semantics \cite{mu2018all, yang2018breaking, gao2018representation, li-etal-2020-sentence}.
Such limitations may prevent PNNs from correctly adopting entity information, reducing the possibility of getting optimal results.

\section{Distance in Prototypical Neural Networks}
\label{sec:pnn_assumption}

In this section, we describe PNNs' feed-forward propagation from the mathematical viewpoint focusing on the PNNs' distance function.
In $K$-way $N$-shot, let $\mathbb{S}_k$ denote the small support set containing $N$ labeled examples with the class $k$.
PNNs calculate the prototype of each class through mean-aggregating the embedded support examples:
\begin{equation}
    \rvc_k = \frac{1}{|\sS_k|}\sum_{(\rvx_i \in \sS_k)} f_\phi(\rvx_i)
\end{equation}
where $f_\phi$ is the embedding network.
The class probabilities of a query data $\mathbf{x}$ are given by a distance function $d$ following a softmax operation:
\begin{equation}
    p_{\phi}(\rvy = k \mid \rvx) = \frac{exp(-d(f_\phi(\rvx), \rvc_k)}{\sum\nolimits_{k'}exp(-d(f_\phi(\rvx), \rvc_{k'})}
\end{equation}

\begin{theorem}
\label{Euclidean distance and Priori Gaussian in PNN}
Assume data embeddings of the support and query set are independent and identically distributed. Let $\rvc_k$ be the class prototype calculated by an aggregation function $proto(\cdot): \prod_{i=1}^{N}\rmH_i \mapsto \rvh \in \rmH$, the problem: $\mathop{min}_{proto(\cdot)} J$
, where $J$ is the classification loss, achieves minimization given by $proto(\cdot)$ being the arithmetic mean.
\end{theorem}
\begin{corollary}
\label{PNN distribution}
Based on the support set, PNNs estimate a Gaussian distribution $\rmN_k(\rvc_k, \sigma^2)$ for the embeddings in the class $k$ ($\sigma$ is a constant vector). 
The corresponding choice of the Bregman divergence $d$ should be the squared Euclidean distance.
\end{corollary}

Proofs are provided in the Appendix~\ref{sec:appendix_pnn}.
While $d$ is proposed to be any Bregman Divergence \cite{banerjee2005clustering, snell2017prototypical}, we prove the optimal distance function should be the squared Euclidean distance: $\|\rvz - \rvz' \|^2$.
\footnote{According to corollary~\ref{PNN distribution}, PNNs require the embeddings to follow Gaussian distribution.
Similarly, works \cite{yang2020free, hu2022squeezing} empirically follow the assumption and propose corresponding embedding post-processions to achieve performance gains.}

PNNs consider that distances between embeddings and prototypes represent the entity probabilities.
Therefore, we count on the distance to capture the sharing entity information between the word and the prototypes.
Factorizing the distance $(-\| f_\phi(\rvx) - \rvc_{k}\|^2)$ to $(-\| f_\phi(\rvx) \|^2+2f_\phi(\rvx)^T\rvc_{k}-\| \rvc_{k} \|^2)$, the entity probabilities are not only proportional to the dot production, but are also reversely proportional to the two $\ell2$-norms.
While $\| f_\phi(\rvx) \|^2$ represents query data information, different $\| \rvc_{k} \|^2$ implies part of the probabilities are priorly determined, and the word is more likely to be classified to the entity that has the smaller prototype $\ell2$-norm.
Unfortunately, because of the representation degeneration of PTMs, these priorly determined probabilities tend to introduce non-entity information, and bias the PNNs' distance towards frequencies.

\section{Representation Degeneration of Pre-trained Models}
\label{rd_ptms}

In this section, we introduce the concept of representation degeneration in PTMs and explain its associated effects to PNNs.
Small-scale PTMs, like GloVe \cite{pennington-etal-2014-glove} and Word2Vec \cite{Mikolov2013EfficientEO}, are argued by researchers as low-capacity models for representing the richness in semantics of natural languages \cite{yang2018breaking, zhao2018softmax}. 
Both theoretical \cite{gao2018representation} and empirical \cite{mu2018all} results in literature have proven:
the learned word embeddings contain substantial non-semantic statistics information, \emph{i.e.} the frequencies of the words, causing a lower performance on various downstream tasks, like the task of word classification \cite{mu2018all}.

\begin{figure}[!ht]
    \centering
    \includegraphics[width=0.48\textwidth]{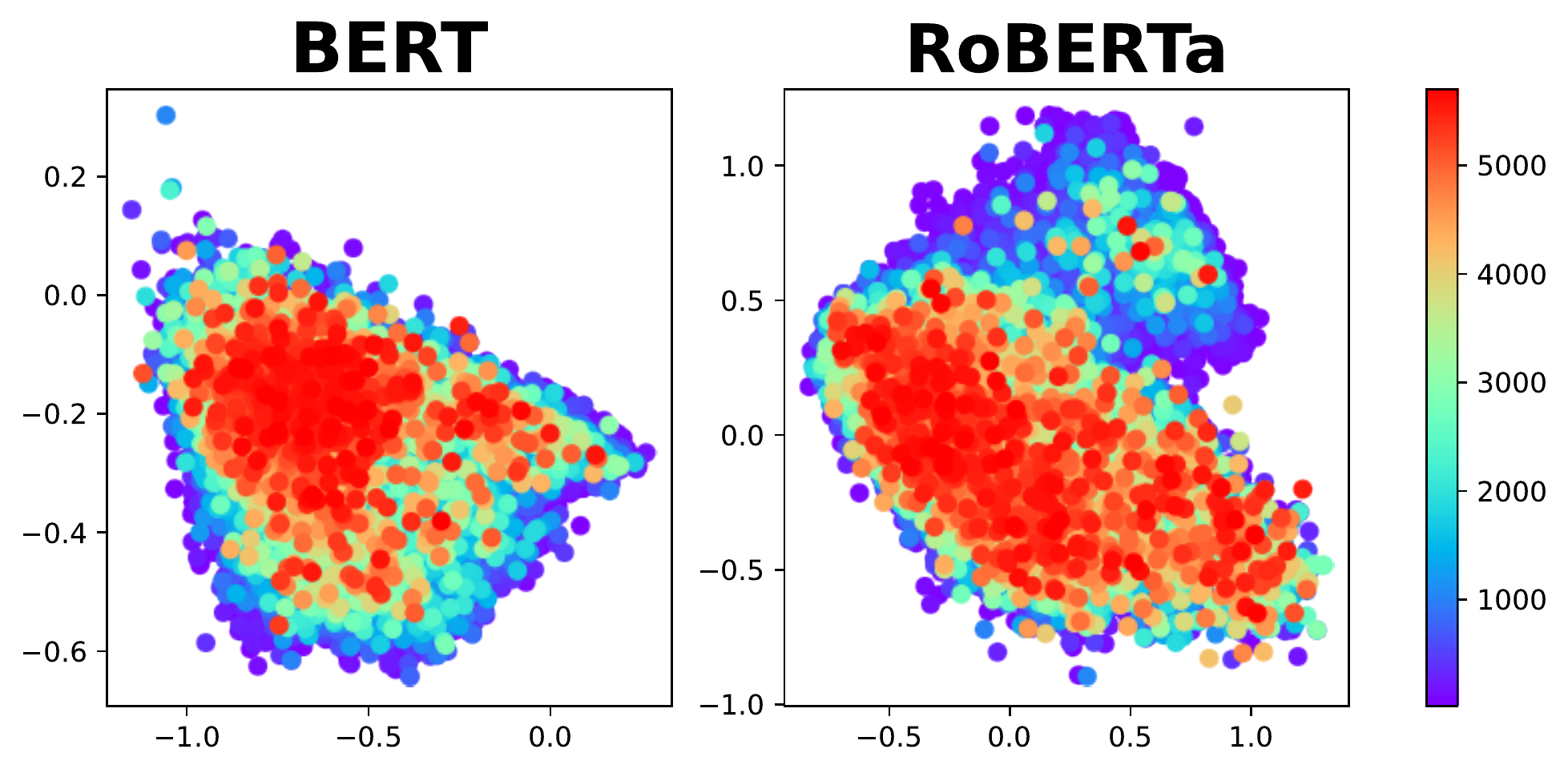}
    \caption{The first two coefficients of PCA analysis on the word embeddings. Color represents frequencies.  The deep colors are clustered.}
    \label{fig:wd_fq}
\end{figure}
Recent Transformer \cite{vaswani2017attention}-based large-scale PTMs \cite{devlin-etal-2019-bert, liu2019roberta} are groundbreaking in modeling natural language. 
However, we are concerned that the learned embeddings might also contain the information regarding the non-semantic word frequencies. 
In line with \cite{mu2018all}, we use the online statistics data\footnote{Data are taken from the Corpus of Contemporary American English (COCA) that provides 60000 English words with frequencies (COCA\_60000).}, get the word embeddings from BERT and RoBERTa, and do principal component analysis (PCA) to extract the first two coefficients, and plot them on point diagrams.
Figure~\ref{fig:wd_fq} displays the result.
The results show both models' embeddings have correlation to frequencies.

In addition, \cite{li-etal-2020-sentence} finds in the embedding space, word embedding $\ell2$-norms are inversely proportional to their frequencies.
As in PNNs, the prototypes are the mean-aggregation of the words in the support set.
Therefore, the prototype $\ell2$-norms are also correlated to word frequencies as well as the priorly determined probabilities we find in section~\ref{sec:pnn_assumption}.
However, we hypothesize that word frequencies are shallow statistics that are irrelevant to word entities, and the priorly determined probabilities represent little entity information.

\begin{figure}[!ht]
    \centering
    \includegraphics[width=0.48\textwidth]{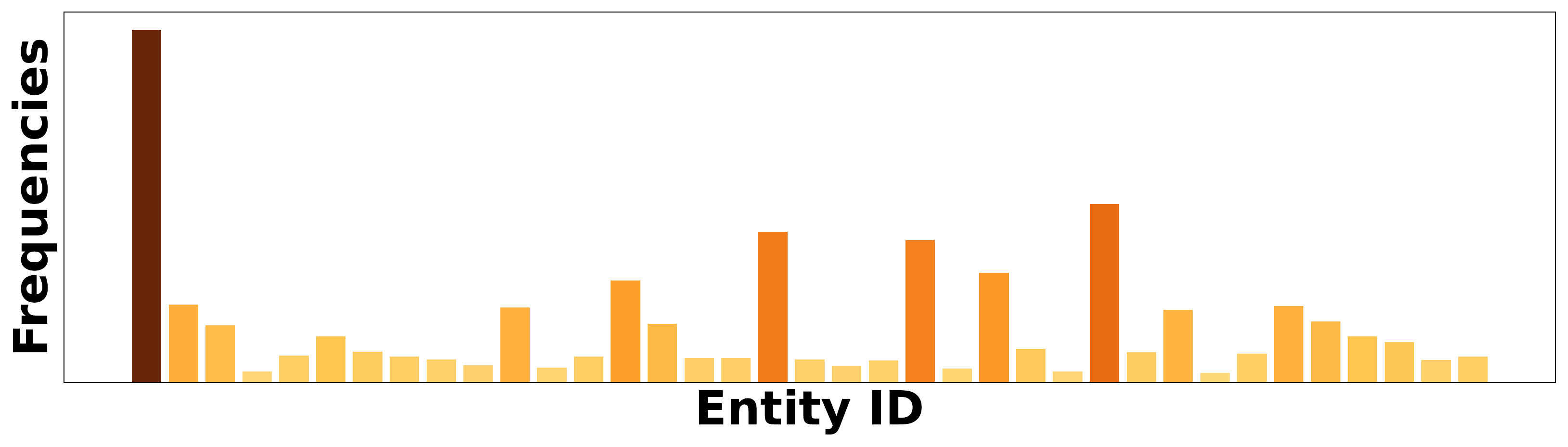}
    \caption{A bar chart displaying the mean word frequencies of different entities. The deeper the color, the larger the mean frequency.}
    \label{fig:wfe}
\end{figure}
\begin{figure}[!ht]
    \centering
    \includegraphics[width=0.48\textwidth]{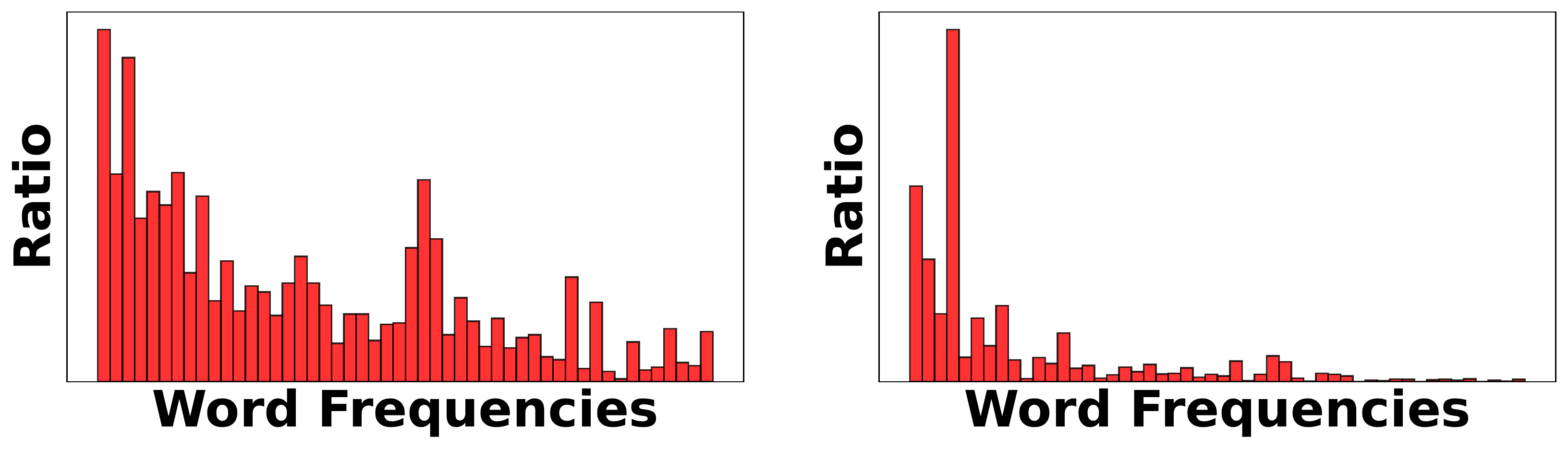}
    \caption{A histogram displaying the word frequencies in two entities.}
    \label{fig:wfw}
\end{figure}
We empirically demonstrate the irrelevance between entities and frequencies in this section.
We will demonstrate the irrelevance between prototype $\ell2$-norms and entities in the next section.
In a few-shot NER dataset \cite{ding-etal-2021-nerd}, we count the mean word frequencies of different entities and the frequencies of each word in two random sampled entities.\footnote{The words frequencies are counted on the first 2.5 million sentences in BookCorpus \cite{zhu2015aligning} processed by HuggingFace \cite{wolf-etal-2020-transformers}.}
Figure~\ref{fig:wfe} and Figure~\ref{fig:wfw} display the results.
Frequencies can be similar among different entities yet distinct in the same entity.
Same as the analysis in the next section, we suppose that this irrelevance introduces non-entity information into PNNs probabilities, and biases the PNNs distance towards focusing on frequencies.

\section{Distance Bias of Prototypical Neural Networks}
\label{sec:pnn_degeneration}

In section~\ref{sec:pnn_assumption}, we have shown that PNNs have a priori on the distances between word embeddings and different entities: embeddings are more likely to be close to the entity that has a smaller prototype $\ell2$-norm, and the word is more likely to be classified to that entity.
However, in section~\ref{rd_ptms}, we argue this priori will introduce non-entity information that confuses the calculation of probabilities in PNNs.
We have shown frequencies and entities are irrelevant.
In the following two figures, we further show the prototype $\ell2$-norms vary in a manner that is also irrelevant to entities.

\begin{figure}[!ht]
    \centering
    \includegraphics[width=0.4\textwidth]{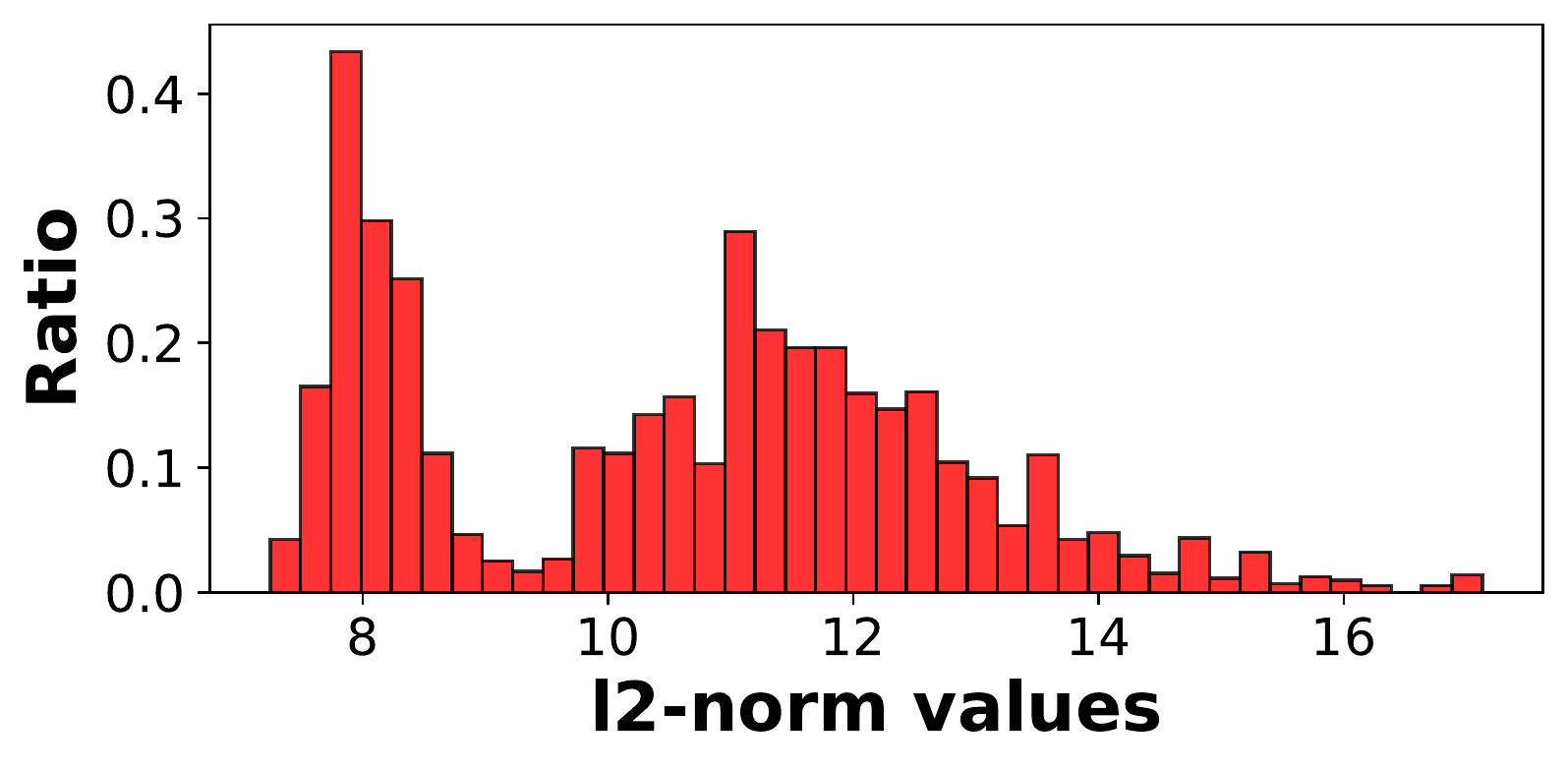}
    \caption{A histogram displays the average prototypes $\ell2$-norm of all classes.}
    \label{fig:all-l2-norm}
\end{figure}

\begin{figure}[!ht]
    \centering
    \includegraphics[width=0.4\textwidth]{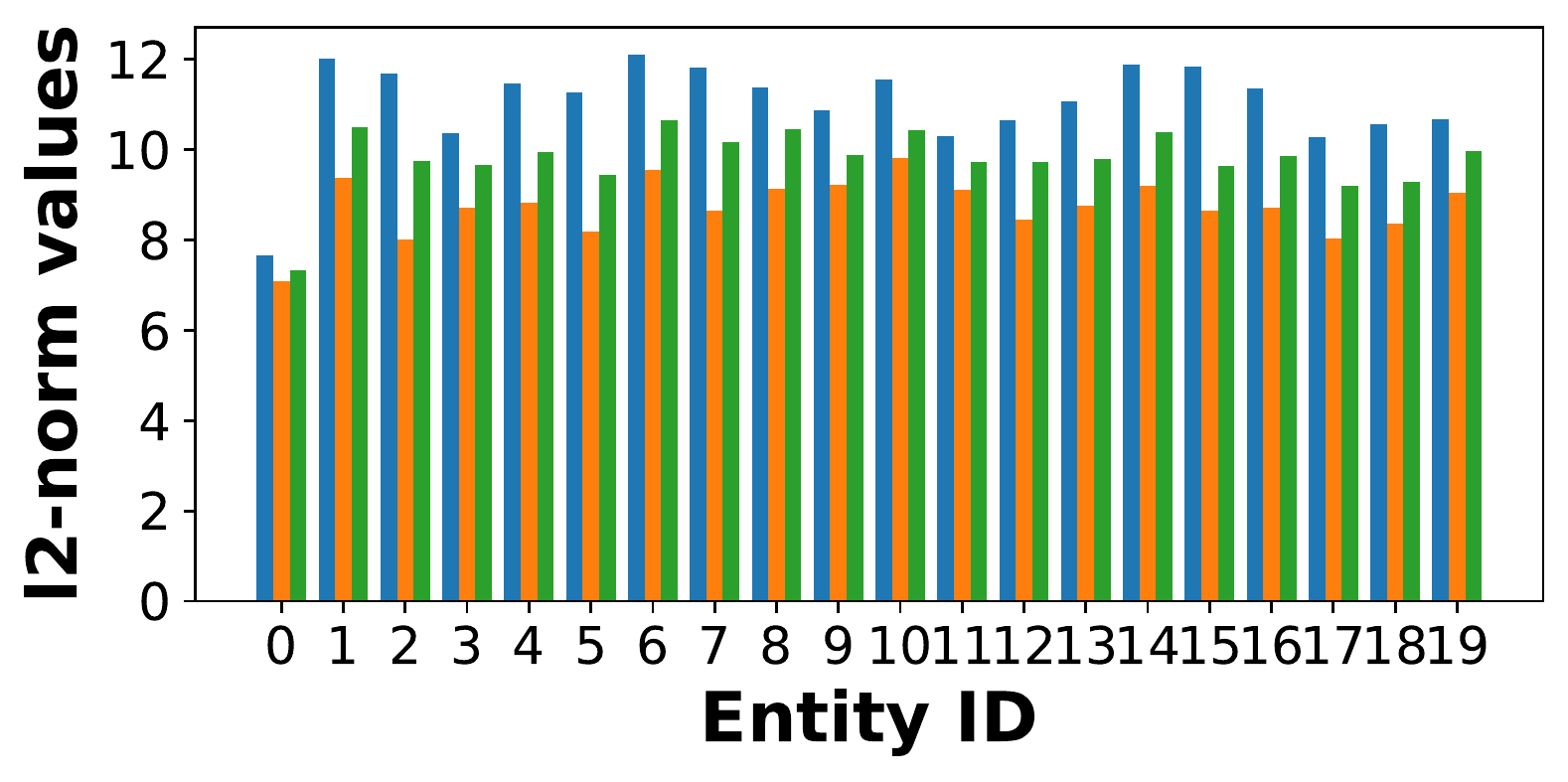}
    \caption{Max(Blue)/Avg.(Green)/Min(Orange) prototypes $\ell2$-norms within a same class.}
    \label{fig:classl2norm}
\end{figure}

Figure~\ref{fig:all-l2-norm} displays the average prototype l2-norms of all classes.
The $\ell2$-norms vary greatly among different classes (min=7.25, max=17.13, coefficient of variation=0.202).
In Figure~\ref{fig:classl2norm}, the blue column represents the largest class-prototype $\ell2$-norm, the orange one the smallest and the green one the average.
Even within the same class, the prototypes $\ell2$-norms demonstrate large variance due to the contrasting difference among episodes.

Distances between prototypes and word embeddings should represent entity probabilities.
Unfortunately, with respect to the above problem, the distances in the original PNNs are biased towards frequencies instead of being entity-oriented.
As a result, PNNs tend to overfit the training data and be trained with unreliable loss minimization.

\subsection{The Overfitting Problem}
\label{overfit}

In this section, we aim to account for the overfitting problem caused by the biased distance.
Let $\sS_u$ be the embeddings of few-labeled data set and $\sQ_u$ be the embeddings of the query data set.

\begin{theorem}
\label{PNN learning 2}
PNNs learn on a Markov Chain: $\sS_u \to \sQ_u$, and maximizes the information bound on the mutual information between $\sS_u$ and $\sQ_u$.
\end{theorem}
\begin{corollary}
\label{PNN overfitting}
Let $\sS_u^g$ be unknown embeddings that the Markov chain: $\sS_u^g \to \sQ_u$ holds according to entity information. The integrated Markove chain becomes: $\sS_u^g \to \sS_u \to \sQ_u$, and PNNs will overfit the words frequencies information in $\sS_u$.
\end{corollary}

Proofs are provided in the Appendix~\ref{sec:appendix_bd}-\ref{sec:appendix_pnn}. 
PNNs learn to maximize the information bound of the mutual information between the support and query data, where the information bound is modeled by the frequency-related distances.
However, it is because frequencies are irrelevant to entities.
Thus, frequency-related distances will confuse PNNs with incorrect evidences, \emph{i.e.} word frequencies, when connecting labeled and query data, preventing PNNs from learning meaningful entity information.
As the frequencies can change randomly on new classes, the distances can no longer correctly model the entity probabilities on a new testing data.

\subsection{Unreliable Empirical Loss Minimization}

In this section, we provide a further explanation to the problem of unreliable empirical loss minimization of training PNNs with biased distances.
Given a hypothesis space $\mathcal{H}$ and its element $h$\footnote{$\mathcal{H}$ can be the all potential parameters of a given network structure and $h$ can be an arbitrary parameter.}, we aim at minimizing the expected loss to find the optimal solution for a given task:
\begin{equation}
    R(h) =  \int \ell(h(\rvx_i, y_i)) d p(\rvx, y)
\end{equation}
Noted that $p(\rvx, y)$ is unknown and we use the empirical loss in practical as a proxy for $R(h)$:
\begin{equation}
    R_I(h) =  \frac{1}{I}\sum_{i=1}^{I} \ell(h(\rvx_i, y_i))
\end{equation}
Let $h^*=\mathop{argmin}_{h\in\mathcal{H}}R(h)$ be the hypothesis that minimizes the expected loss and $h_I=\mathop{argmin}_{h\in\mathcal{H}}R_I(h)$ be the hypothesis that minimizes the empirical loss. 
The approximation error $[R(h_I) - R(h^*)]$ quantifies the degree of closeness to the optimal result $h_I$.
Noting that the frequency information guides the loss minimization during training PNNs as analyzed in section~\ref{overfit}.
Due to the uncertainty of word frequencies, a good approximation on the training data can have a large approximation error on the testing, which can jeopardize PNNs testing performance.

Moreover, the labeled examples for each episode are limited to $N$-shot, where data in each episode is not likely to cover many words.
As such, the frequencies of the words and prototype $\ell2$-norms can vary among episodes, resulting in unstable training with low efficiency in model learning and lowering the testing performance.

\section{Normalizing the Prototypes}
\label{sec:n_pnn}

In this section, we aim to provide a solution to the above-mentioned problems through a normalizing method.
Varying $\ell2$-norms mainly causes frequency-biased distances and the above two problems.
As a result, we consider normalizing the prototypes to $\ell2$-norm-invariant vectors. 
Earlier works in Computer Vision find normalizing both prototypes and the query data embeddings can achieve better and more stable results \cite{gidaris2018dynamic}. 
However, we do not normalize the query data embeddings, because word embeddings represent more detail and other useful information that may be eliminated by the normalization.

Representing high-level entity information, prototypes should not be priorly distinguished from each other.
Furthermore, observing the following evidence, we argue that prototype $\ell2$-norms have limited contribution to the correct classification.

\begin{figure}[!ht]
    \centering
    \includegraphics[width=0.4\textwidth]{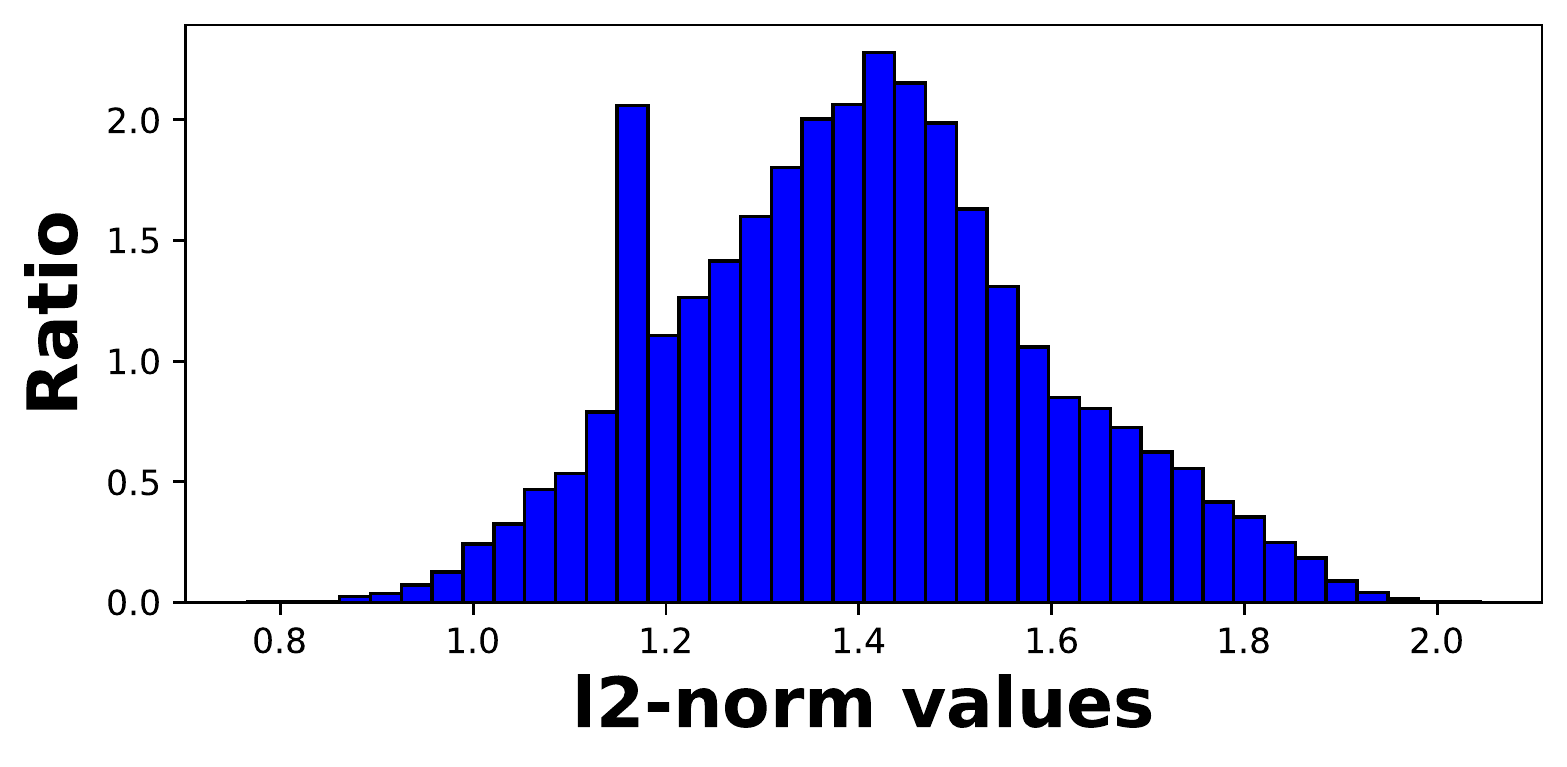}
    \caption{A histogram displays the $\ell2$-norms of the pre-trained classifier in BERT.}
    \label{fig:p_cls_norm}
\end{figure}
\begin{figure}[!ht]
    \centering
    \includegraphics[width=0.4\textwidth]{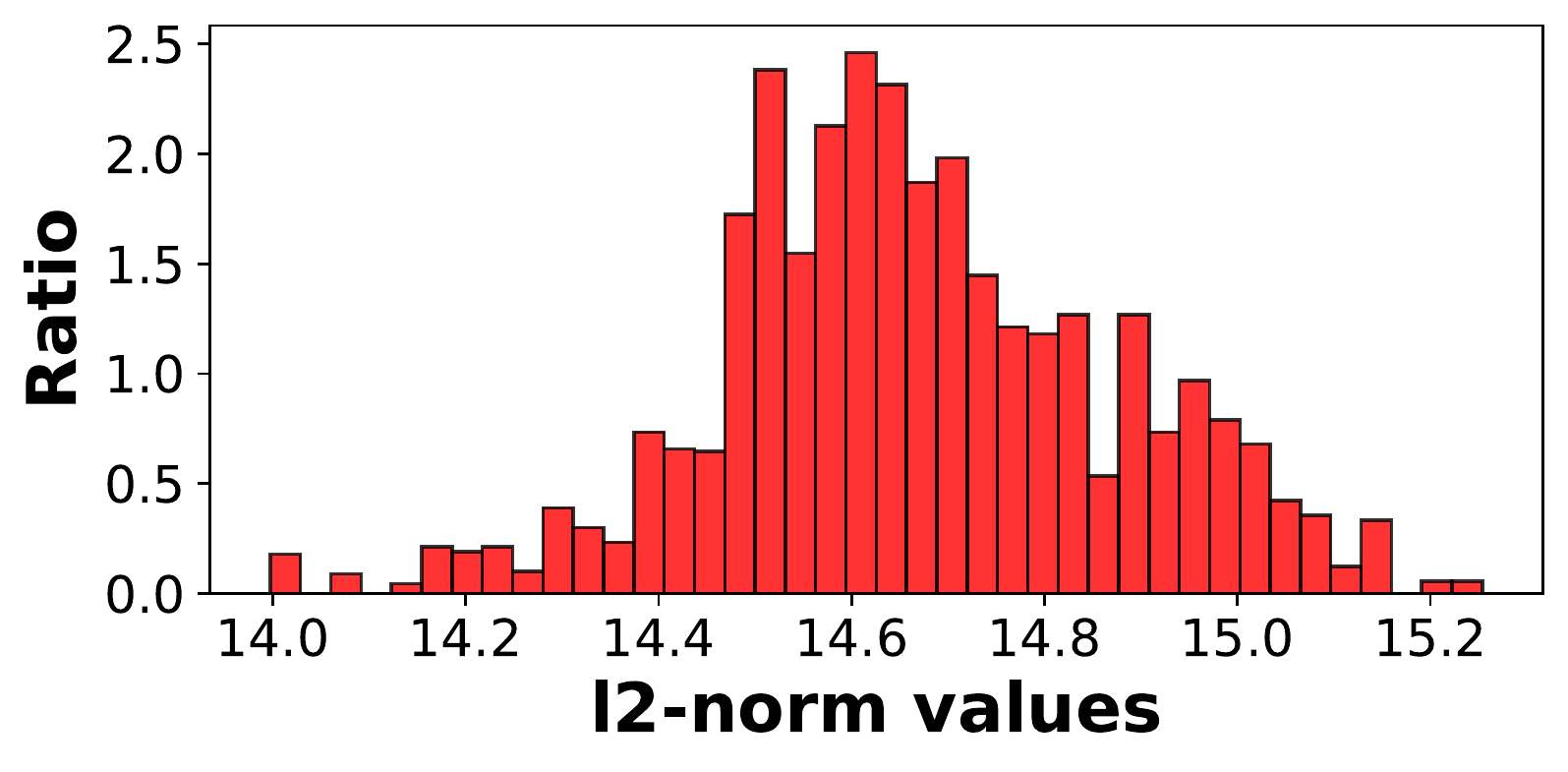}
    \caption{A histogram displays the average prototypes $\ell2$-norms of all classes after training.}
    \label{fig:pnn_trained_norm}
\end{figure}
In both the BERT's pre-training (1) and the original PNNs (2), we find the $\ell2$-norms of class features play limited roles to the correct classification.
\begin{itemize}
    \item[(1)]
    In the BERT's pre-training that predicts a word by its context, the $\ell2$-norms of the words features, \emph{i.e.} rows of the classifier, show subtle variance.
    Figure~\ref{fig:p_cls_norm} presents the $\ell2$-norms of the classifier rows: min=0.766, max=2.045, coefficient of variation=0.138. 
    \item[(2)]
    Without any intervention to the original PNNs, after the training, the prototype $\ell2$-norms vary much less compared to the original, \emph{i.e.} after the training: (min=14.00, max=15.25, coefficient of variation=0.014) compared to the original :(min=7.25, max=17.13, coefficient of variation=0.202), and Figure~\ref{fig:pnn_trained_norm} compared to Figure~\ref{fig:all-l2-norm}.
\end{itemize}

Based on the above analysis, we propose to normalize the prototypes to unit vectors before calculating the class probabilities.
\begin{algorithm}[!ht]
   \caption{Normalizing the Prototypes}
   \label{alg:PN normalized algorithm}
\begin{algorithmic}
   \STATE {\textcolor{brown}{*** Pseudo-code in PyTorch ***}}
   \STATE {import torch.nn.functional as F}
   \STATE {$\bm{C}$ = Calculate Prototype ($\mathbb{S}$) $\in \mathbb{R}^{k \times h}$}
   \STATE {\textcolor{brown}{*** The Normalization ***}}
   \STATE {\textbf{$\bm{C}$ = F.normalize ($\bm{C}$, dim=-1})}
   \STATE {... the same as the original PNNs ...}
\end{algorithmic}
\end{algorithm}

\noindent \textbf{Connection to the Adaptive Loss:} Different data may associate with different difficulties to be classified. 
Adaptive loss is proposed to be able to change dynamically in magnitude so as to capture the difficult ones \cite{han-etal-2021-exploring, oreshkin2018tadam, li2020boosting}.
Humans are prone to processing high-frequency words as reported in psychological studies \cite{brysbaert2018word}.
Applying this psychological finding to the named entity recognition in natural language processing, we postulate that if a word appears more frequently, its entity should be easier to be classified.
To this end, PNNs well adapt to task difficulty through the frequency-related embedding $\ell2$-norms of the query data.

\section{Experiments \& Results}

To demonstrate the effectiveness of our normalized PNNs, we conduct experiments on nine few-shot named entity recognition datasets proposed by \cite{huang-etal-2021-shot} and \cite{ding-etal-2021-nerd}.

\noindent \textbf{Datasets:} Being a classical and basic natural language understanding task, dozens of supervised NER datasets have been proposed, including WikiGold \cite{balasuriya2009named}, CoNLL 2003 \cite{ sang2003introduction}, WNUT 2017 \cite{derczynski2017results}, MIT Movie \cite{liu2013query}, MIT Restaurant \cite{liu2013asgard}, SNIPS \cite{coucke2018snips}, ATIS \cite{hakkani2016multi}, Multiwoz \cite{budzianowski2018multiwoz}.
Based on these datasets, researchers \cite{huang-etal-2021-shot} re-structure them to the "$K$-way $N$-shot" few-shot setting into a comprehensive few-shot NER benchmark.
However, except for the formatting change of data, the simple and direct re-structuring shall lose track of some critical NER properties, such as the task-difficulty differences between the fine-grained and coarse-grained entities \cite{ding-etal-2021-nerd}.
Therefore, a new expert and challenging dataset has been proposed as a benchmark in few-shot NER \cite{ding-etal-2021-nerd}.

\noindent \textbf{Experimental Settings:} Without special notations, we basically follow the original implementations in the two open sources\footnote{\url{https://github.com/thunlp/Few-NERD}}\footnote{\url{https://github.com/few-shot-NER-benchmark}}, including models, training/testing pipelines, hyper-parameters, and sampled episodes.
We report results using the standard evaluation metrics: micro averaged F1 score.
We re-run all the experiments of the origin PNNs to examine the performance improvements by our normalization method based on the same hardware device. We add early stop constraints when reproducing results of \cite{huang-etal-2021-shot}) and relocate the comparable results from the peer models \cite{das-etal-2022-container, ding-etal-2021-nerd}.
\footnote{The replicated performances are inferior to the reported results in the related works, so we use the reported results for a standard reference.}
All the experiments are conducted on a single 3090Ti GPU.

\begin{table*}[htbp]
    \centering
    \small
    \caption{The performance State-of-the-art models and our method on FEW-NERD.}
    \begin{threeparttable}
    \begin{tabular}{{lccccc}}
        \toprule
        \multirow{2}{*}{\textbf{Model}} & \multicolumn{4}{c}{\textbf{FEW-NERD(INTRA) F1 scores}}  & \multirow{2}{*}{\textbf{Avg.}}\\
        \cmidrule(lr){2-2}\cmidrule(lr){3-3}\cmidrule(lr){4-4}\cmidrule(lr){5-5}
        & \textbf{5 way 1\textasciitilde2 shot} & \textbf{5 way 5\textasciitilde10 shot} & \textbf{10 way 1\textasciitilde2 shot} & \textbf{10 way 5\textasciitilde10 shot}\\
        \midrule
        Struct (EMNLP 2020) & 30.21 & 38.00 & 21.03 & 26.42 & 28.92 \\
        NNShot (EMNLP 2020) & 25.75 & 36.18 & 18.27 & 27.38 & 26.90 \\
        CONTaiNER (ACL 2020) & \textbf{40.43} & 53.70 & \textbf{33.84} & 47.49 & \textbf{43.87} \\
        \hspace{3mm}+Viterbi (ACL 2020)  & \textbf{40.43} & 53.71 & 33.82 & 47.51 & 43.86 \\
        Proto (Neurips 2017) & 20.76 & 42.54 & 15.05 & 35.40 & 28.43 \\
        \rowcolor{maroon!10}
        Proto$_{\mathrm{ours}}$* & 36.83 & \textbf{54.62} & 30.06 & \textbf{47.61} & 42.28 \\
        \midrule
        \multirow{2}{*}{\textbf{Model}} & \multicolumn{4}{c}{\textbf{FEW-NERD(INTER) F1 scores}}  & \multirow{2}{*}{\textbf{Avg.}}\\
        \cmidrule(lr){2-2}\cmidrule(lr){3-3}\cmidrule(lr){4-4}\cmidrule(lr){5-5}
        & \textbf{5 way 1\textasciitilde2 shot} & \textbf{5 way 5\textasciitilde10 shot} & \textbf{10 way 1\textasciitilde2 shot} & \textbf{10 way 5\textasciitilde10 shot}\\
        \midrule
        Struct (EMNLP 2020) & 51.88 & 57.32 & 43.34 & 49.57 & 50.53 \\
        NNShot (EMNLP 2020) & 47.24 & 55.64 & 38.87 & 49.57 & 47.83 \\
        CONTaiNER (ACL 2020) & 55.95 & 61.83 & 48.35 & 57.12 & 55.81 \\
        \hspace{3mm}+Viterbi (ACL 2020)  & \textbf{56.10} & 61.90 & \textbf{48.36} & 57.13 & 55.87 \\
        Proto (Neurips 2017) & 38.83 & 58.79 & 32.34 & 52.92 & 45.72 \\
        \rowcolor{maroon!10}
        Proto$_{\mathrm{ours}}$* & 54.35 & \textbf{66.93} & 47.32 & \textbf{61.50} & \textbf{57.52} \\
        \bottomrule
    \end{tabular}
    \begin{tablenotes}    
        \footnotesize               
        \item[*] We change the learning rate from 1e-4 to 1e-5. We lower the learning rate because normalized PNN converges too rapidly to be tested on dev set (given the same evaluation steps) before it overfits the training set. 
    \end{tablenotes}
    \end{threeparttable}
    \label{tab:few-nerd}
\end{table*}
\noindent \textbf{Comparison to the State-Of-The-Art Methods:} 
We compare the normalized PNN (Proto$_\mathrm{ours}$) to four advanced methods on Few-NERD.
"Struct" and "NNShot" are proposed by \cite{yang-katiyar-2020-simple}.
"NNShot" classifies the query data to its nearest data entity in the embedding space, and "Struct" further leverages the Viterbi decoding \cite{1450960} to produce the final results.
"CONTaiNER" as well as the Viterbi enhanced version are proposed by \cite{das-etal-2022-container}.
It utilizes contrastive learning to differentiate word entities.
And unlike PNN, "NNShot" and "Struct", "CONTaiNER" will be fine-tuned on the new entities using the limited labeled examples.

We briefly introduce the main characteristic of Few-NERD:
it defines entity types from two perspectives called the fine-grained (INTER) and coarse-grained (INTRA).
Under the fine-grained definition, different entities can share more abstract similarities.
For example, entities "Island" and "Mountain" are both "Location", and entities "Director" and "Athlete" are both "Person".
Under the coarse-grained definition, entities have more differences, such as "Location" v.s. "Person" and "Event" v.s. "Organization".
If the training classes contain "Island", the model can easily identify the entity "Mountain" at the testing because they share the same "Location" information.
Therefore, training on the fine-grained set is less challenging for NER on new testing entities.

Table~\ref{tab:few-nerd} reports our normalized PNNs on Few-NERD as well as the results of state-of-the-art models, and the original PNNs for comparisons.
Compared with the original PNNs, the normalization achieves at least 8.14\% performance gain (largest: 16.07\% and average: 12.82\%).
The sophisticated contrastive learning-based CONTaiNER outperforms our method in certain settings.
On average, our model is slightly superior (49.84\% (Proto$_\mathrm{ours}$) v.s. 49.80\%).
Besides, CONTaiNER needs to be fine-tuned on the testing data in order to alleviate the differences between training and testing entities, which can account for its superior performance on the coarse-grained (INTRA) set.
It should be noted that our normalization method shows competitive performances yet maintains the PNNs' advantages, \emph{i.e.} the low computation cost and easy implementation.
In addition, our model achieves the highest average F1 scores (57.52\% (Proto$_\mathrm{ours}$)) on the fine-grained (INTER) set, demonstrating its superiority in a more practical setting \cite{ding-etal-2021-nerd}.

\begin{table*}[htbp]
    \caption{The performance on benchmark datasets proposed by \cite{huang-etal-2021-shot}.}
    \centering
    \scriptsize
    \begin{threeparttable}
    \begin{tabular}{{lccccccccc}}
        \toprule
        \multirow{2}{*}{\textbf{Model}} & \multicolumn{8}{c}{\textbf{Datasets (5-shot) F1 scors}}  & \multirow{2}{*}{\textbf{Avg.}}\\
        \cmidrule(lr){2-2}\cmidrule(lr){3-3}\cmidrule(lr){4-4}\cmidrule(lr){5-5}
        \cmidrule(lr){6-6}\cmidrule(lr){7-7}\cmidrule(lr){8-8}\cmidrule(lr){9-9}
        & \textbf{CoNLL} & \textbf{WikiGold} & \textbf{WNUT17} & \textbf{MIT Movie} & \textbf{MIT Restaurant} & \textbf{SNIPS} & \textbf{ATIS} & \textbf{Multiwoz}\\
        \midrule
        Proto* & 58.22 & 47.58 & 20.51 & 29.94 & 43.65 & 56.96 & 73.82 & 23.74 & 44.30 \\
        \rowcolor{maroon!10}
        Proto$_\mathrm{ours}$* & 58.70 & 55.69 & 28.46 & 50.01 & 51.34 & 76.69 & 87.41 & 27.78 & 54.51 \\
        Proto+NSP* & 62.92 & 63.33 & 33.87 & 35.25 & 44.15 & 51.66 & 74.58 & 40.52 & 50.79 \\
        \rowcolor{maroon!10}
        Proto$_\mathrm{ours}$+NSP* & \textbf{66.50} & 67.63 & \textbf{37.75} & 51.32 & \textbf{54.98} & \textbf{83.17} & \textbf{90.47} & \textbf{47.26} & \textbf{62.39}\\
        LC+NSP+ST** & 65.4 & \textbf{68.4} & 37.6 & \textbf{55.9} & 51.3 & 83.0 & \textbf{90.5} & 45.1 & 62.12\\
        \bottomrule
    \end{tabular}
    \begin{tablenotes}    
        \footnotesize               
        \item[*] For meaningful comparison and to calculate the performance gains, we re-run the baseline models "Proto" and "Proto$_\mathrm{ours}$" with the same setting. To reduce the time cost, we add the early stop constraints, i.e. stop the training if a continual 5 epochs training does not improve the dev-set F1 scores.
        \item[**] The replicated results in [*] are lower than the reported results in the original paper. Therefore, we directly copy the results in the original paper as a comparison for demonstrating our method's effectiveness.
    \end{tablenotes}
    \end{threeparttable}
    \label{tab:ner benchmarks}
\end{table*}
\noindent \textbf{Incorporation with the Data-Driven Pre-training:}
\cite{huang-etal-2021-shot} proposes two pre-training techniques called noisy supervised pre-training (NSP) and self-training (ST).
NSP utilizes the large-scale noisy labeled entities on Wikipedia to pre-train the models, while ST utilizes an NER system (teacher model) to label large-scale unlabeled datasets to pre-train the student models.
Both the techniques seek extra supervisions to help the model tackle the challenges of the few-shot classification.
\cite{huang-etal-2021-shot} chooses two baselines: the linear classification (LC) and PNNs. 
And on ten re-structured few-shot NER datasets, they compare the performances of the two baselines as well as the two baselines plus the two pre-training techniques. They report the best performance is achieved by the combination of "LC+NSP+ST".

Because the processed datasets "I2B2" and "Onto" are not open-sourced by \cite{huang-etal-2021-shot}, we conduct the experiments on the other eight datasets.
For more details of the datasets, please refer to \cite{huang-etal-2021-shot}.

Table~\ref{tab:ner benchmarks} reports the results on the eight datasets.
Results vary among different datasets, but the normalized PNNs consistently outperform the original PNNs (min:0.48\%, max:20.07\%, average:10.20\%). 
In certain datasets, normalized PNNs achieves extremely close even higher results than the original PNNs plus a pre-training method that is expensive in time cost (Proto+NSP).
Furthermore, higher performance gains are obtained when incorporating the normalized PNNs with the NSP technique (Proto+NSP +6.48\% v.s. Proto$_\mathrm{ours}$+NSP +7.88\%).
Our results show that the classical PNNs combined with the simple normalization and NSP can achieve the best results on the eight few-shot NER datasets (the open sources do not provide the ST checkpoints for PNN).
This finding is innovative compared to the results in \cite{huang-etal-2021-shot}.

\noindent \textbf{Effective Learning:}
Figure~\ref{fig.learningdraw} (in Appendix~\ref{sec:appendix_elas}) visualizes the training and dev F1 scores on two settings of Few-NERD, including the original and the normalized PNNs (the * mark denotes that we set the learning rate to $1e^-5$ as the same as our experimental settings).
Comparing the red with blue lines (with the same learning rate), normalized PNNs can fit the training data in a faster mode yet can achieve higher Dev F1 scores.
Comparing the red with green lines, setting the learning rate to $1e^{-4}$ and without normalizing, PNNs learn unstably and more significantly overfit the training data (in INTER 5 way 5\textasciitilde5 shot, dev F1 scores decreases before increasing, and in INTRA 10 way 1\textasciitilde2 shot, the increasing of training F1 scores results in decreasing of dev F1 scores).

\noindent \textbf{Ablation Studies:}
Based on our analysis in section~\ref{sec:n_pnn}, we only normalize the prototypes and leave the query data embeddings unchanged.
We conduct ablation studies about the normalization strategies on Few-NERD as shown in Table~\ref{tab:few-nerd-ab} in Appendix~\ref{sec:appendix_elas}.
Proto$_{AB1}$ means we normalize only the query data embeddings and leave the prototypes unchanged, and Proto$_{AB2}$ means we normalize both the prototypes and the query data embeddings.
We provide four sub-cases for ablation studies.
All cases report substantial performance decrease.

\section{Conclusion}

We examine the synergistic effects of the large-scale PTMs and the classical PNNs in the few-shot NER.
Our theoretical analysis of PNNs shows that PNNs' distances that represent the query data's entity probabilities are partly priorily determined in terms of the prototype $\ell2$-norms.
However, on the embeddings of the PTMs, we empirically verify that embedding $\ell2$-norms contain little entity information, being a type of PTMs' representation degeneration.
Furthermore, we show that such representation degeneration makes PNNs' distance biased towards frequencies instead of entity-denoting.
This distance bias prevents PNNs from learning useful entity information and causes PNNs to overfit the training corpus and become unreliable on new entities.
Therefore, We propose a one-line-code normalization remedy to reconcile PTMs and PNNs for few-shot NER.
The experimental results based on nine datasets suggest that the normalized PNNs proposed in this work achieve significant performance improvements over the original PNNs and get competitive results compared with the latest sophisticated methods while maintaining PNNs' all advantages, such as easy implementation and low computation cost.
Considering the promising results and the innovation in normalizing the existing models, our results and analysis may be an interest of reference study for researchers and practitioners working with few-shot NER or other relevant tasks that involve the use of PTMs or PNNs.

\section{Limitations}

There are certain limitations in this paper. While our theoretical analysis about PNNs and the concept of PTMs' representation degeneration are not limited to the few-shot named-entity recognition, our focused problem, \emph{e.g.} PNNs' distance is biased towards frequencies, is based on the fact that the greatly varied word frequencies represent limited entity information.
It is possible that in other tasks, the corpus frequencies can represent semantic features, or the frequencies change much less.
Our normalization remedy, therefore, cannot be directly applied to those tasks.
Also, representation degeneration is a crucial intrinsic problem of large-scale PTMs.
Our focused aspects, \emph{e.g.} frequencies and entity information, is one type of practical issue.
We argue that such intrinsic problems can result in different practical issues affecting other NLP tasks beyond this current work's scope.

\section*{Acknowledgement}

This work is supported by National Key R\&D Program of China (2020AAA0105200). 

\bibliography{main}

\begin{thebibliography}{48}
\expandafter\ifx\csname natexlab\endcsname\relax\def\natexlab#1{#1}\fi

\bibitem[{Balasuriya et~al.(2009)Balasuriya, Ringland, Nothman, Murphy, and
  Curran}]{balasuriya2009named}
Dominic Balasuriya, Nicky Ringland, Joel Nothman, Tara Murphy, and James~R
  Curran. 2009.
\newblock Named entity recognition in wikipedia.
\newblock In \emph{Proceedings of the 2009 workshop on the people’s web meets
  NLP: Collaboratively constructed semantic resources (People’s Web)}, pages
  10--18.

\bibitem[{Banerjee et~al.(2005)Banerjee, Merugu, Dhillon, Ghosh, and
  Lafferty}]{banerjee2005clustering}
Arindam Banerjee, Srujana Merugu, Inderjit~S Dhillon, Joydeep Ghosh, and John
  Lafferty. 2005.
\newblock Clustering with bregman divergences.
\newblock \emph{Journal of machine learning research}, 6(10).

\bibitem[{Bao et~al.(2020)Bao, Wu, Chang, and Barzilay}]{bao2020fewshot}
Yujia Bao, Menghua Wu, Shiyu Chang, and Regina Barzilay. 2020.
\newblock Few-shot text classification with distributional signatures.
\newblock In \emph{International Conference on Learning Representations}.

\bibitem[{Brysbaert et~al.(2018)Brysbaert, Mandera, and
  Keuleers}]{brysbaert2018word}
Marc Brysbaert, Pawe{\l} Mandera, and Emmanuel Keuleers. 2018.
\newblock The word frequency effect in word processing: An updated review.
\newblock \emph{Current Directions in Psychological Science}, 27(1):45--50.

\bibitem[{Budzianowski et~al.(2018)Budzianowski, Wen, Tseng, Casanueva, Ultes,
  Ramadan, and Gasic}]{budzianowski2018multiwoz}
Pawe{\l} Budzianowski, Tsung-Hsien Wen, Bo-Hsiang Tseng, I{\~n}igo Casanueva,
  Stefan Ultes, Osman Ramadan, and Milica Gasic. 2018.
\newblock Multiwoz-a large-scale multi-domain wizard-of-oz dataset for
  task-oriented dialogue modelling.
\newblock In \emph{Proceedings of the 2018 Conference on Empirical Methods in
  Natural Language Processing}, pages 5016--5026.

\bibitem[{Coucke et~al.(2018)Coucke, Saade, Ball, Bluche, Caulier, Leroy,
  Doumouro, Gisselbrecht, Caltagirone, Lavril et~al.}]{coucke2018snips}
Alice Coucke, Alaa Saade, Adrien Ball, Th{\'e}odore Bluche, Alexandre Caulier,
  David Leroy, Cl{\'e}ment Doumouro, Thibault Gisselbrecht, Francesco
  Caltagirone, Thibaut Lavril, et~al. 2018.
\newblock Snips voice platform: an embedded spoken language understanding
  system for private-by-design voice interfaces.
\newblock \emph{arXiv preprint arXiv:1805.10190}.

\bibitem[{Das et~al.(2022)Das, Katiyar, Passonneau, and
  Zhang}]{das-etal-2022-container}
Sarkar Snigdha~Sarathi Das, Arzoo Katiyar, Rebecca Passonneau, and Rui Zhang.
  2022.
\newblock \href {https://aclanthology.org/2022.acl-long.439} {{CONT}ai{NER}:
  Few-shot named entity recognition via contrastive learning}.
\newblock In \emph{Proceedings of the 60th Annual Meeting of the Association
  for Computational Linguistics (Volume 1: Long Papers)}, pages 6338--6353,
  Dublin, Ireland. Association for Computational Linguistics.

\bibitem[{Derczynski et~al.(2017)Derczynski, Nichols, van Erp, and
  Limsopatham}]{derczynski2017results}
Leon Derczynski, Eric Nichols, Marieke van Erp, and Nut Limsopatham. 2017.
\newblock Results of the wnut2017 shared task on novel and emerging entity
  recognition.
\newblock In \emph{Proceedings of the 3rd Workshop on Noisy User-generated
  Text}, pages 140--147.

\bibitem[{Devlin et~al.(2019)Devlin, Chang, Lee, and
  Toutanova}]{devlin-etal-2019-bert}
Jacob Devlin, Ming-Wei Chang, Kenton Lee, and Kristina Toutanova. 2019.
\newblock \href {https://doi.org/10.18653/v1/N19-1423} {{BERT}: Pre-training of
  deep bidirectional transformers for language understanding}.
\newblock In \emph{Proceedings of the 2019 Conference of the North {A}merican
  Chapter of the Association for Computational Linguistics: Human Language
  Technologies, Volume 1 (Long and Short Papers)}, pages 4171--4186,
  Minneapolis, Minnesota. Association for Computational Linguistics.

\bibitem[{Ding et~al.(2021)Ding, Xu, Chen, Wang, Han, Xie, Zheng, and
  Liu}]{ding-etal-2021-nerd}
Ning Ding, Guangwei Xu, Yulin Chen, Xiaobin Wang, Xu~Han, Pengjun Xie, Haitao
  Zheng, and Zhiyuan Liu. 2021.
\newblock \href {https://doi.org/10.18653/v1/2021.acl-long.248} {Few-{NERD}: A
  few-shot named entity recognition dataset}.
\newblock In \emph{Proceedings of the 59th Annual Meeting of the Association
  for Computational Linguistics and the 11th International Joint Conference on
  Natural Language Processing (Volume 1: Long Papers)}, pages 3198--3213,
  Online. Association for Computational Linguistics.

\bibitem[{Forney(1973)}]{1450960}
G.D. Forney. 1973.
\newblock \href {https://doi.org/10.1109/PROC.1973.9030} {The viterbi
  algorithm}.
\newblock \emph{Proceedings of the IEEE}, 61(3):268--278.

\bibitem[{Gao et~al.(2018)Gao, He, Tan, Qin, Wang, and
  Liu}]{gao2018representation}
Jun Gao, Di~He, Xu~Tan, Tao Qin, Liwei Wang, and Tieyan Liu. 2018.
\newblock Representation degeneration problem in training natural language
  generation models.
\newblock In \emph{International Conference on Learning Representations}.

\bibitem[{Gidaris and Komodakis(2018)}]{gidaris2018dynamic}
Spyros Gidaris and Nikos Komodakis. 2018.
\newblock Dynamic few-shot visual learning without forgetting.
\newblock In \emph{Proceedings of the IEEE conference on computer vision and
  pattern recognition}, pages 4367--4375.

\bibitem[{Hakkani-T{\"u}r et~al.(2016)Hakkani-T{\"u}r, T{\"u}r, Celikyilmaz,
  Chen, Gao, Deng, and Wang}]{hakkani2016multi}
Dilek Hakkani-T{\"u}r, G{\"o}khan T{\"u}r, Asli Celikyilmaz, Yun-Nung Chen,
  Jianfeng Gao, Li~Deng, and Ye-Yi Wang. 2016.
\newblock Multi-domain joint semantic frame parsing using bi-directional
  rnn-lstm.
\newblock In \emph{Interspeech}, pages 715--719.

\bibitem[{Han et~al.(2021)Han, Cheng, and Lu}]{han-etal-2021-exploring}
Jiale Han, Bo~Cheng, and Wei Lu. 2021.
\newblock \href {https://doi.org/10.18653/v1/2021.emnlp-main.204} {Exploring
  task difficulty for few-shot relation extraction}.
\newblock In \emph{Proceedings of the 2021 Conference on Empirical Methods in
  Natural Language Processing}, pages 2605--2616, Online and Punta Cana,
  Dominican Republic. Association for Computational Linguistics.

\bibitem[{Hochreiter et~al.(2001)Hochreiter, Younger, and
  Conwell}]{hochreiter2001learning}
Sepp Hochreiter, A~Steven Younger, and Peter~R Conwell. 2001.
\newblock Learning to learn using gradient descent.
\newblock In \emph{International Conference on Artificial Neural Networks},
  pages 87--94. Springer.

\bibitem[{Holla et~al.(2020)Holla, Mishra, Yannakoudakis, and
  Shutova}]{holla2020learning}
Nithin Holla, Pushkar Mishra, Helen Yannakoudakis, and Ekaterina Shutova. 2020.
\newblock Learning to learn to disambiguate: Meta-learning for few-shot word
  sense disambiguation.
\newblock In \emph{Findings of the Association for Computational Linguistics:
  EMNLP 2020}, pages 4517--4533.

\bibitem[{Hu et~al.(2022)Hu, Pateux, and Gripon}]{hu2022squeezing}
Yuqing Hu, St{\'e}phane Pateux, and Vincent Gripon. 2022.
\newblock Squeezing backbone feature distributions to the max for efficient
  few-shot learning.
\newblock \emph{Algorithms}, 15(5):147.

\bibitem[{Huang et~al.(2021)Huang, Li, Subudhi, Jose, Balakrishnan, Chen, Peng,
  Gao, and Han}]{huang-etal-2021-shot}
Jiaxin Huang, Chunyuan Li, Krishan Subudhi, Damien Jose, Shobana Balakrishnan,
  Weizhu Chen, Baolin Peng, Jianfeng Gao, and Jiawei Han. 2021.
\newblock \href {https://doi.org/10.18653/v1/2021.emnlp-main.813} {Few-shot
  named entity recognition: An empirical baseline study}.
\newblock In \emph{Proceedings of the 2021 Conference on Empirical Methods in
  Natural Language Processing}, pages 10408--10423.

\bibitem[{Koch et~al.(2015)}]{koch2015siamese}
Gregory Koch et~al. 2015.
\newblock Siamese neural networks for one-shot image recognition.

\bibitem[{Lample et~al.(2016)Lample, Ballesteros, Subramanian, Kawakami, and
  Dyer}]{lample-etal-2016-neural}
Guillaume Lample, Miguel Ballesteros, Sandeep Subramanian, Kazuya Kawakami, and
  Chris Dyer. 2016.
\newblock \href {https://doi.org/10.18653/v1/N16-1030} {Neural architectures
  for named entity recognition}.
\newblock In \emph{Proceedings of the 2016 Conference of the North {A}merican
  Chapter of the Association for Computational Linguistics: Human Language
  Technologies}, pages 260--270, San Diego, California. Association for
  Computational Linguistics.

\bibitem[{Li et~al.(2020{\natexlab{a}})Li, Huang, Lan, Feng, Li, and
  Wang}]{li2020boosting}
Aoxue Li, Weiran Huang, Xu~Lan, Jiashi Feng, Zhenguo Li, and Liwei Wang.
  2020{\natexlab{a}}.
\newblock Boosting few-shot learning with adaptive margin loss.
\newblock In \emph{Proceedings of the IEEE/CVF conference on computer vision
  and pattern recognition}, pages 12576--12584.

\bibitem[{Li et~al.(2020{\natexlab{b}})Li, Zhou, He, Wang, Yang, and
  Li}]{li-etal-2020-sentence}
Bohan Li, Hao Zhou, Junxian He, Mingxuan Wang, Yiming Yang, and Lei Li.
  2020{\natexlab{b}}.
\newblock \href {https://doi.org/10.18653/v1/2020.emnlp-main.733} {On the
  sentence embeddings from pre-trained language models}.
\newblock In \emph{Proceedings of the 2020 Conference on Empirical Methods in
  Natural Language Processing (EMNLP)}, pages 9119--9130, Online. Association
  for Computational Linguistics.

\bibitem[{Liu et~al.(2013{\natexlab{a}})Liu, Pasupat, Cyphers, and
  Glass}]{liu2013asgard}
Jingjing Liu, Panupong Pasupat, Scott Cyphers, and Jim Glass.
  2013{\natexlab{a}}.
\newblock Asgard: A portable architecture for multilingual dialogue systems.
\newblock In \emph{2013 IEEE International Conference on Acoustics, Speech and
  Signal Processing}, pages 8386--8390. IEEE.

\bibitem[{Liu et~al.(2013{\natexlab{b}})Liu, Pasupat, Wang, Cyphers, and
  Glass}]{liu2013query}
Jingjing Liu, Panupong Pasupat, Yining Wang, Scott Cyphers, and Jim Glass.
  2013{\natexlab{b}}.
\newblock Query understanding enhanced by hierarchical parsing structures.
\newblock In \emph{2013 IEEE Workshop on Automatic Speech Recognition and
  Understanding}, pages 72--77. IEEE.

\bibitem[{Liu et~al.(2019)Liu, Ott, Goyal, Du, Joshi, Chen, Levy, Lewis,
  Zettlemoyer, and Stoyanov}]{liu2019roberta}
Yinhan Liu, Myle Ott, Naman Goyal, Jingfei Du, Mandar Joshi, Danqi Chen, Omer
  Levy, Mike Lewis, Luke Zettlemoyer, and Veselin Stoyanov. 2019.
\newblock Roberta: A robustly optimized bert pretraining approach.
\newblock \emph{arXiv preprint arXiv:1907.11692}.

\bibitem[{Mikolov et~al.(2013)Mikolov, Chen, Corrado, and
  Dean}]{Mikolov2013EfficientEO}
Tomas Mikolov, Kai Chen, Gregory~S. Corrado, and Jeffrey Dean. 2013.
\newblock Efficient estimation of word representations in vector space.
\newblock In \emph{ICLR}.

\bibitem[{Mu and Viswanath(2018)}]{mu2018all}
Jiaqi Mu and Pramod Viswanath. 2018.
\newblock All-but-the-top: Simple and effective postprocessing for word
  representations.
\newblock In \emph{International Conference on Learning Representations}.

\bibitem[{Oreshkin et~al.(2018)Oreshkin, Rodr{\'\i}guez~L{\'o}pez, and
  Lacoste}]{oreshkin2018tadam}
Boris Oreshkin, Pau Rodr{\'\i}guez~L{\'o}pez, and Alexandre Lacoste. 2018.
\newblock Tadam: Task dependent adaptive metric for improved few-shot learning.
\newblock \emph{Advances in neural information processing systems}, 31.

\bibitem[{Pennington et~al.(2014)Pennington, Socher, and
  Manning}]{pennington-etal-2014-glove}
Jeffrey Pennington, Richard Socher, and Christopher Manning. 2014.
\newblock \href {https://doi.org/10.3115/v1/D14-1162} {{G}lo{V}e: Global
  vectors for word representation}.
\newblock In \emph{Proceedings of the 2014 Conference on Empirical Methods in
  Natural Language Processing ({EMNLP})}, pages 1532--1543, Doha, Qatar.
  Association for Computational Linguistics.

\bibitem[{Sang and De~Meulder(2003)}]{sang2003introduction}
Erik Tjong~Kim Sang and Fien De~Meulder. 2003.
\newblock Introduction to the conll-2003 shared task: Language-independent
  named entity recognition.
\newblock In \emph{Proceedings of the Seventh Conference on Natural Language
  Learning at HLT-NAACL 2003}, pages 142--147.

\bibitem[{Snell et~al.(2017)Snell, Swersky, and Zemel}]{snell2017prototypical}
Jake Snell, Kevin Swersky, and Richard Zemel. 2017.
\newblock Prototypical networks for few-shot learning.
\newblock \emph{Advances in neural information processing systems}, 30.

\bibitem[{Stubbs and Uzuner(2015)}]{stubbs2015annotating}
Amber Stubbs and {\"O}zlem Uzuner. 2015.
\newblock Annotating longitudinal clinical narratives for de-identification:
  The 2014 i2b2/uthealth corpus.
\newblock \emph{Journal of biomedical informatics}, 58:S20--S29.

\bibitem[{Sung et~al.(2018)Sung, Yang, Zhang, Xiang, Torr, and
  Hospedales}]{sung2018learning}
Flood Sung, Yongxin Yang, Li~Zhang, Tao Xiang, Philip~HS Torr, and Timothy~M
  Hospedales. 2018.
\newblock Learning to compare: Relation network for few-shot learning.
\newblock In \emph{Proceedings of the IEEE conference on computer vision and
  pattern recognition}, pages 1199--1208.

\bibitem[{Tong et~al.(2021)Tong, Wang, Xu, Cao, Liu, Hou, and
  Li}]{tong-etal-2021-learning}
Meihan Tong, Shuai Wang, Bin Xu, Yixin Cao, Minghui Liu, Lei Hou, and Juanzi
  Li. 2021.
\newblock \href {https://doi.org/10.18653/v1/2021.acl-long.487} {Learning from
  miscellaneous other-class words for few-shot named entity recognition}.
\newblock In \emph{Proceedings of the 59th Annual Meeting of the Association
  for Computational Linguistics and the 11th International Joint Conference on
  Natural Language Processing (Volume 1: Long Papers)}, pages 6236--6247,
  Online. Association for Computational Linguistics.

\bibitem[{Van~den Oord et~al.(2018)Van~den Oord, Li, and
  Vinyals}]{van2018representation}
Aaron Van~den Oord, Yazhe Li, and Oriol Vinyals. 2018.
\newblock Representation learning with contrastive predictive coding.
\newblock \emph{arXiv e-prints}, pages arXiv--1807.

\bibitem[{Vaswani et~al.(2017)Vaswani, Shazeer, Parmar, Uszkoreit, Jones,
  Gomez, Kaiser, and Polosukhin}]{vaswani2017attention}
Ashish Vaswani, Noam Shazeer, Niki Parmar, Jakob Uszkoreit, Llion Jones,
  Aidan~N Gomez, {\L}ukasz Kaiser, and Illia Polosukhin. 2017.
\newblock Attention is all you need.
\newblock \emph{Advances in neural information processing systems}, 30.

\bibitem[{Vinyals et~al.(2016)Vinyals, Blundell, Lillicrap, Wierstra
  et~al.}]{vinyals2016matching}
Oriol Vinyals, Charles Blundell, Timothy Lillicrap, Daan Wierstra, et~al. 2016.
\newblock Matching networks for one shot learning.
\newblock \emph{Advances in neural information processing systems}, 29.

\bibitem[{Wang et~al.(2018)Wang, Feng, Chen, Yu, Huang, and
  Yu}]{wang2018visual}
Jindong Wang, Wenjie Feng, Yiqiang Chen, Han Yu, Meiyu Huang, and Philip~S Yu.
  2018.
\newblock Visual domain adaptation with manifold embedded distribution
  alignment.
\newblock In \emph{Proceedings of the 26th ACM international conference on
  Multimedia}, pages 402--410.

\bibitem[{Wang et~al.(2020)Wang, Yao, Kwok, and Ni}]{wang2020generalizing}
Yaqing Wang, Quanming Yao, James~T Kwok, and Lionel~M Ni. 2020.
\newblock Generalizing from a few examples: A survey on few-shot learning.
\newblock \emph{ACM computing surveys (csur)}, 53(3):1--34.

\bibitem[{Wilson and Cook(2020)}]{wilson2020survey}
Garrett Wilson and Diane~J Cook. 2020.
\newblock A survey of unsupervised deep domain adaptation.
\newblock \emph{ACM Transactions on Intelligent Systems and Technology (TIST)},
  11(5):1--46.

\bibitem[{Wolf et~al.(2020)Wolf, Debut, Sanh, Chaumond, Delangue, Moi, Cistac,
  Rault, Louf, Funtowicz, Davison, Shleifer, von Platen, Ma, Jernite, Plu, Xu,
  Scao, Gugger, Drame, Lhoest, and Rush}]{wolf-etal-2020-transformers}
Thomas Wolf, Lysandre Debut, Victor Sanh, Julien Chaumond, Clement Delangue,
  Anthony Moi, Pierric Cistac, Tim Rault, Rémi Louf, Morgan Funtowicz, Joe
  Davison, Sam Shleifer, Patrick von Platen, Clara Ma, Yacine Jernite, Julien
  Plu, Canwen Xu, Teven~Le Scao, Sylvain Gugger, Mariama Drame, Quentin Lhoest,
  and Alexander~M. Rush. 2020.
\newblock \href {https://www.aclweb.org/anthology/2020.emnlp-demos.6}
  {Transformers: State-of-the-art natural language processing}.
\newblock In \emph{Proceedings of the 2020 Conference on Empirical Methods in
  Natural Language Processing: System Demonstrations}, pages 38--45.

\bibitem[{Yadav and Bethard(2019)}]{yadav2019survey}
Vikas Yadav and Steven Bethard. 2019.
\newblock A survey on recent advances in named entity recognition from deep
  learning models.
\newblock \emph{arXiv preprint arXiv:1910.11470}.

\bibitem[{Yang et~al.(2020)Yang, Liu, and Xu}]{yang2020free}
Shuo Yang, Lu~Liu, and Min Xu. 2020.
\newblock Free lunch for few-shot learning: Distribution calibration.
\newblock In \emph{International Conference on Learning Representations}.

\bibitem[{Yang and Katiyar(2020)}]{yang-katiyar-2020-simple}
Yi~Yang and Arzoo Katiyar. 2020.
\newblock \href {https://doi.org/10.18653/v1/2020.emnlp-main.516} {Simple and
  effective few-shot named entity recognition with structured nearest neighbor
  learning}.
\newblock In \emph{Proceedings of the 2020 Conference on Empirical Methods in
  Natural Language Processing (EMNLP)}, pages 6365--6375.

\bibitem[{Yang et~al.(2018)Yang, Dai, Salakhutdinov, and
  Cohen}]{yang2018breaking}
Zhilin Yang, Zihang Dai, Ruslan Salakhutdinov, and William~W Cohen. 2018.
\newblock Breaking the softmax bottleneck: A high-rank rnn language model.
\newblock In \emph{International Conference on Learning Representations}.

\bibitem[{Zhao et~al.(2018)Zhao, Zhao, Wan, and Zhang}]{zhao2018softmax}
Yue Zhao, Deli Zhao, Shaohua Wan, and Bo~Zhang. 2018.
\newblock Softmax supervision with isotropic normalization.

\bibitem[{Zhu et~al.(2015)Zhu, Kiros, Zemel, Salakhutdinov, Urtasun, Torralba,
  and Fidler}]{zhu2015aligning}
Yukun Zhu, Ryan Kiros, Rich Zemel, Ruslan Salakhutdinov, Raquel Urtasun,
  Antonio Torralba, and Sanja Fidler. 2015.
\newblock Aligning books and movies: Towards story-like visual explanations by
  watching movies and reading books.
\newblock In \emph{Proceedings of the IEEE international conference on computer
  vision}, pages 19--27.

\end{thebibliography}
\bibliographystyle{acl_natbib}

\clearpage
\appendix
\section{Bregman Divergence}
\label{sec:appendix_bd}

\begin{definition}[Bregman (1967); Censor and Zenios (1998)]
\label{bd_0}
Let $\phi: \mathcal{S} \mapsto \mathbb{R}$, $\mathcal{S} = dom(\phi)$ be a strictly
convex function defined on a convex set $\mathcal{S} \subseteq \mathbb{R}^d$ such that $\phi$ is differentiable on $ri(S)$, assumed to
be nonempty. The Bregman divergence $d_\phi: \mathcal{S} \times ri(\mathcal{S}) \mapsto [0,\infty)$ is defined as
\begin{equation}
    d_\phi = \phi(x) - \phi(y) - \langle x - y, \nabla \phi(y) \rangle
\end{equation}
where $\nabla \phi(y)$ represents the gradient vector of $\phi$ evaluated at $y$.
\end{definition}

\begin{proposition}[Banerjee (2005)]
\label{bd_1}
Let $X$ be a random variable that take values in $\mathcal{X} = \{x_i\}_{i=1}^n \subset \mathcal{S} \subseteq \mathcal{R}^d$ following a positive probability measure $\nu$ such that $E_\nu[X] \in ri(\mathcal{S})$. Given a Bregman divergence $d_\phi: \mathcal{S} \times ri(\mathcal{S}) \mapsto [0, \infty)$, the problem
\begin{equation}
    \mathop{min}_{s\in ri(\mathcal{S})} E_\nu[d_\phi(X, s)]
\end{equation}
has a unique minimizer given by $s^\dag = \mu = E_\nu[X]$.
\end{proposition}

\begin{theorem}[Banerjee (2005)]
\label{bd_2}
Let $p_{(\psi, \theta)}$ be the probability density function of a regular exponential family distribution. Let $\phi$ be the conjugate function of $\psi$ so that $(int(dom(\phi)),\phi)$ is the Legendre dual of $(\Theta,\Psi)$. Let $\theta \in \Theta$ be the natural parameter and $\mu \in int(dom(\phi)) $be the corresponding expectation parameter.
Let $d_\phi$ be the Bregman divergence derived from $\phi$. Then $p_{(\psi,\theta)}$ can be uniquely expressed as
\begin{equation}
    p_{(\psi,\theta)}(x) = exp(-d_\phi(x, \mu))b_\phi(x), \quad \forall x \in dom(\phi)
\end{equation}
where $b_\phi: dom(\phi) \mapsto \mathbb{R}_+$ is a uniquely determined function.
\end{theorem}

\clearpage
\section{Prototypical Neural Networks}
\label{sec:appendix_pnn}

\begin{algorithm}[ht]
   \caption{\textit{K}-way \textit{N}-shot Prototypical Neural Network}
   \label{alg:PN raw algorithm}
\begin{algorithmic}
   \STATE {\bfseries Input:} An episode $E_i$ containing: support data $\sS_u$ and query data $\sQ_u$.
   \STATE {\bfseries Output:} The loss $J$ for the episode $E_i$.
   \STATE \#Calculating prototypes on $\sS_u$
   \STATE $C$ = NewEmptyList(Length=K)
   \FOR{$k=1$ {\bfseries to} $K$}
   \STATE $\rvc_k$ = $\frac{1}{N_k}\sum_{(\rvx^{i}, \rvy^{i}==k)}f_{enc}(\rvx^{i})$
   \ENDFOR
   \STATE \#Classification on $\sQ_u$ and calculating the loss $J$
   \STATE $J$ = NewEmptyList(Length=0)
   \FOR{$k=1$ {\bfseries to} $K$}
   \FOR{$(\rvx^{i}, \rvy^{i}==k)$ in  $\sQ_u$}
   \STATE $Q(\hat{\rvy}^{i}==k|\rvx^{i})=\frac{exp(-d_\phi(f_{enc}(\rvx^{i}), \rvc_k))}
   {\sum\nolimits_{k'=1}^{K}exp(-d_\phi(f_{enc}(\rvx^{i}), \rvc_{k'}))}$
   \STATE $J$.Add(CrossEntropyLoss($\hat{\rvy}^{i}$, $\rvy^{i}$))
   \ENDFOR
   \ENDFOR
   \STATE $J$ = Mean($J$)
\end{algorithmic}
\end{algorithm}

\begin{remark}
\label{s and q}
Prototype calculation and query data classification are independent but have the same goal of minimizing the classifying loss.
\end{remark}

\begin{theorem_}
Assume data embeddings of the support and query data are independent and identically distributed. Let $\rvc_k$ be the class prototype calculated by an aggregation function $proto(\cdot): \prod_{i=1}^{N}\rmH_i \mapsto \rvh \in \rmH$, the problem
\begin{equation*}
    \mathop{min}_{proto(\cdot)} J
\end{equation*}
,where $J$ is the classifying loss, achieves minimization given by $proto(\cdot)$ being the arithmetic mean.
\end{theorem_}

\begin{proof}

In the above Remark, we argue the prototype calculation should also minimize the classifying loss while the query data is unseen.
As the optimal prototypes should minimize the classification loss on query data, and the support and query data are independent and identically distributed, we let the support data be the agency of the query data.
Therefore, the optimal prototype should minimize the classification loss on support data.

Let us consider the $m^{th}$ class, the corresponding cross-entropy loss is:
\begin{equation}
\begin{aligned}
   J_m &= -\sum\limits_i\mathrm{log}\frac{exp(-d_\phi(f_{enc}(\rvx^{i}), \rvc_m))}
   {\sum\nolimits_{k'=1}^{K}exp(-d_\phi(f_{enc}(\rvx^{i}), \rvc_{k'}))}\\
   &=-\sum\limits_i[-d_\phi(f_{enc}(\rvx^i), \rvc_m)\\&-\mathrm{log}\sum\limits_{k'=1}^{K}exp(-d_\phi(f_{enc}(\rvx^i, \rvc_{k'})))]\\
   &=\sum\limits_i d_\phi(f_{enc}(\rvx^i), \rvc_m)\\&+\sum\limits_i\mathrm{log}\sum\limits_{k'=1}^{K}exp(-d_\phi(f_{enc}(\rvx^i, \rvc_{k'})))
\end{aligned}
\end{equation}
where $\rvx^i$ is the support data with the class $m$, $\rvc_m$ and $\rvx_{k'}$ be the $m^{th}$ and $k'^{th}$ class prototype. As we aim to find the optimal $\rvc_m$, we take the derivative of $J_m$ respect to $\rvc_m$:
\begin{equation}
\label{cm_1}
\begin{aligned}
\frac{\partial J_m}{\partial \rvc_m} &= 
\frac{\partial \sum\nolimits_i d_\phi(\rvh^i, \rvc_m)}{\partial \rvc_m}
\\&+\frac{\partial\sum\nolimits_i\mathrm{log}\sum\nolimits_{k'}exp(-d_\phi(\rvh^i, \rvc_{k'}))}{\partial \rvc_m}\\
&=\frac{\partial \sum\nolimits_i d_\phi(\rvh^i, \rvc_m)}{\partial \rvc_m}
\\&+\sum\nolimits_i \frac{{\partial(-d_\phi(\rvh^i, \rvc_m))}/{\partial \rvc_m}}{\sum\nolimits_{k'}exp(-d_\phi(\rvh^i, \rvc_{k'}))}\\
&=\sum\nolimits_i(1-\frac{1}{\sum\nolimits_{k'}exp(-d_\phi(\rvh^i, \rvc_{k'}))}) \\
&\qquad \times {\partial(d_\phi(\rvh^i, \rvc_m))}/{\partial \rvc_m}
\end{aligned}
\end{equation}
where $\rvh^i=f_{enc}(\rvx^i)$. As $d_\phi$ is a Bregman Divergence, according to Proposition~\ref{bd_1}, we have $\frac{\partial E_\nu [d_\phi(\mathcal{H}, s)]}{\partial s}=0$ if and only if $s=E_\nu[\mathcal{H}]$. If we use $\alpha$ to normalize the weight of Equation~\ref{cm_1} to have $\sum\nolimits_i\alpha(1-\frac{exp(-d_\phi(\rvh^i, \rvc_{m}))}{\sum\nolimits_{k'}exp(-d_\phi(\rvh^i, \rvc_{k'}))})=1$, then the optimized $\rvc_m$ can be calculated as:
\begin{equation}
\label{cm_2}
    \rvc_m = \sum\nolimits_i\alpha(1-\frac{1}{\sum\nolimits_{k'}exp(-d_\phi(\rvh^i, \rvc_{k'}))})\rvh^i
\end{equation}
The Equation~\ref{cm_2} show the optimized $\rvc_m$ should be the arithmetic mean of the support data embeddings minus the category confidences.
But the category confidences correspond to the probability normalization of \textit{Softmax}.
If we ignore this, the optimal prototype calculation is the arithmetic mean.

\end{proof}

\begin{corollary_}
Based on the support data, PNNs estimate a Gaussian distribution $\rmN_k(\rvc_k, \sigma^2)$ for embeddings in class $k$, where $\sigma$ is a constant vector. 
And the corresponding choice of the Bregman divergence $d$ should be the squared Euclidean distance.
\end{corollary_}

\begin{proof}

According to \cite{banerjee2005clustering}, for the d-dimension spherical Gaussian distribution, the parameter formula is:
\begin{gather}
    p(\rvx;\theta) = \frac{1}{\sqrt{(2\pi\sigma^2)}}exp(-\frac{\|\rvx-a\|^2}{2\sigma^2}) \\
    \mu = a \\
    \phi(\mu) = \frac{1}{2\sigma^2}\mu \\
    d_\phi(\rvx,\mu) = \frac{1}{2\sigma^2}\|\rvx-\mu\|^2
\end{gather}
The $\mu$ in PNNs is the prototypes, i.e. the arithmetic mean of sampled observations, and it exactly estimates the parameter in Gaussian distribution.
Therefore, the optimal prototype calculation results in estimating a Gaussian distribution for each class.
On a Gaussian distribution where $\sigma$ is a constant, $d_\phi$ corresponds to the squared Euclidean distance.

\end{proof}

\begin{theorem_}
PNNs learn on a Markov Chain: $\sS_u \to \sQ_u$, and maximizes the information bound on the mutual information between $\sS_u$ and $\sQ_u$.
\end{theorem_}

\begin{proof}

According to the Theorem~\ref{bd_2}, a Bregman divergence and a Distribution are connected:
\begin{equation}
\label{bd_3}
    \mathrm{log}(P_{(\psi, \theta)}(\rvh)) = -d_\phi(\rvh, \mu) + \phi(\rvh) + log(p_0(\rvh))
\end{equation}
when $P_{(\psi, \theta)}$ is the Gaussian distribution, we have $\phi(\rvh) = \frac{1}{2\sigma^2}\| \rvh \|_2$ and $p_0$ is uniquely determined.

PNNs calculate the distance between $\rvh$ and $\mu$, which can be viewed as the probability of observing $\rvh$ given $\mu$. This relationship between the support and query data implies the Markov Chain: $\sS_u \to \sQ_u$, for observing the query data is dependent on the support data.

In the right of Equation~\ref{bd_3}, $-d_\phi(\rvh, \mu)$ can be viewed as the probability of observing $\rvh$ given $\mu$, and the rest $\phi(\rvh) + log(p_0(\rvh))$ can be viewed as the probability of observing $\rvh$ unknown $\mu$: $p(\rvh)$.
The first term $p(\rvh \mid \mu)$ is inversely proportional to $\| \rvh \|^2$, while the second $p(\rvh)$ is proportional to $\| \rvh \|^2$.
PNNs maximize $-d_\phi(\rvh, \mu)$, resulting in the implicit minimizing of $p(\rvh)$.
Integratedly, the learnt probability $P_{(\psi, \theta)}(\rvh)$ is proportional to $\frac{p(\rvh \mid \mu)}{p(\rvh)}$.
Substitute this back to the loss:

\begin{gather}
\label{pnn_prove}
\begin{aligned}
    J &= -\mathop{\mathbb{E}}\limits_\mathcal{H}\mathrm{log}\left[ 
    \frac{\frac{p(\rvh^k \mid \mu^k)}{p(\rvh^k)}}{\frac{p(\rvh^k \mid \mu^k)}{p(\rvh^k)} + \sum\nolimits_{k' \neq k}\frac{p(\rvh^k \mid \mu^{k'})}{p(\rvh^{k})}}
    \right]\\
    &=\mathop{\mathbb{E}}\limits_\mathcal{H}\mathrm{log}\left[ 
    1 + \frac{p(\rvh^k)}{p(\rvh^k \mid \mu^k)}\sum_{k' \neq k}\frac{p(\rvh^k \mid \mu^{k'})}{p(\rvh^{k})}
    \right]\\
    &\approx\mathop{\mathbb{E}}\limits_\mathcal{H}\mathrm{log}\\
    & \left[ 
    1 + \frac{p(\rvh^k)}{p(\rvh^k \mid \mu^k)}(K-1)\mathop{\mathbb{E}}\limits_{\mu^{k'}}\frac{p(\rvh^k \mid \mu^{k'})}{p(\rvh^{k})}
    \right]\\
    &=\mathop{\mathbb{E}}\limits_\mathcal{H}\mathrm{log}\left[ 
    1 + \frac{p(\rvh^k)}{p(\rvh^k \mid \mu^k)}(K-1)
    \right]\\
    &\geq=\mathop{\mathbb{E}}\limits_\mathcal{H}\mathrm{log}\left[ 
    \frac{p(\rvh^k)}{p(\rvh^k \mid \mu^k)}K
    \right]\\
    &=-I(\rvh^k, \mu^k) + \mathrm{log}(K)
\end{aligned}
\end{gather}
The results show $I(\rvh^k, \mu^k) \geq \mathrm{log}(K) - J$, which means that PNNs minimize the classification loss to maximize the information bound on the mutual information between $
rvh$ and $\mu$, and integratedly, between the support and query data.
We notice the above detail proof follows the same mathematical process in the works on contrastive learning \cite{van2018representation}

\end{proof}

\begin{corollary_}
Let $\sS_u^g$ be unknown embeddings that the Markov chain: $\sS_u^g \to \sQ_u$ holds according to entity information. The integrated Markove chain becomes: $\sS_u^g \to \sS_u \to \sQ_u$, and PNNs will overfit the words frequencies information in $\sS_u$.
\end{corollary_}

\begin{proof}
In the Markov Chain: $\sS_u^g \to \sS_u \to \sQ_u$, using the data processing inequality, \footnote{http://www.scholarpedia.org/article/Mutual\_information} we have:

\begin{equation}
    I(\sS_u, \sQ_u) \geq I(\sS_u^g, \sQ_u)
\end{equation}

The learnt extra information $I(\sS_u, \sQ_u) - I(\sS_u^g, \sQ_u) \geq 0$ represents PNN's overfitting to $\sS_u$'s words frequencies introduced by the frequency-related distances.

\end{proof}

\clearpage
\section{Effective Learning and Ablation Studies}
\label{sec:appendix_elas}

\begin{figure}[!htbp] 
\centering 
\includegraphics[width=0.7\textwidth]{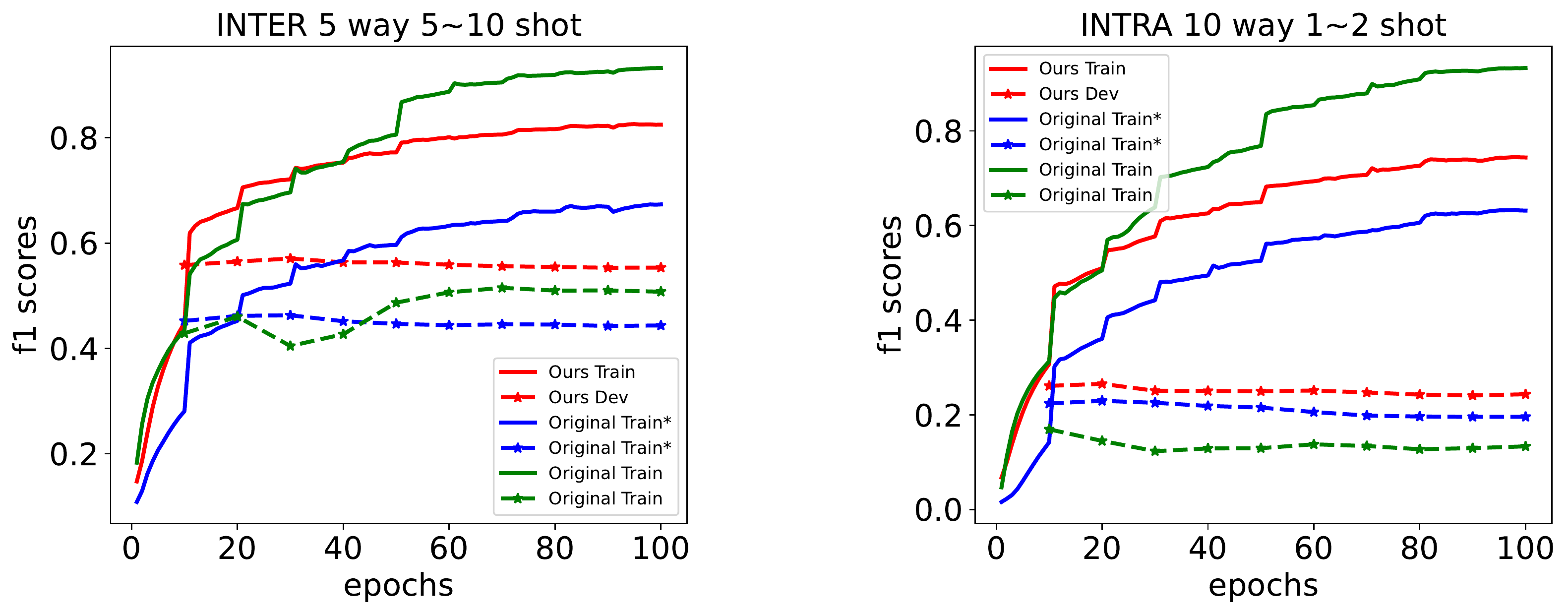} 
\caption{Training and Dev F1 Scores on Few-NERD of two cases.} 
\label{fig.learningdraw} 
\end{figure}

\begin{table}[!htb]
    \centering
    \small
    \caption{Ablation studies of our method on FEW-NERD.}
    \begin{threeparttable}
    \begin{tabular}{{lccccc}}
        \toprule
        \multirow{2}{*}{\textbf{Model}} & \multicolumn{4}{c}{\textbf{FEW-NERD(INTRA) F1 scores}}\\
        \cmidrule(lr){2-2}\cmidrule(lr){3-3}\cmidrule(lr){4-4}\cmidrule(lr){5-5}
        & \textbf{5 way 1\textasciitilde2 shot} & \textbf{5 way 5\textasciitilde10 shot} & \textbf{10 way 1\textasciitilde2 shot} & \textbf{10 way 5\textasciitilde10 shot}\\
        \midrule
        Proto$_\mathrm{ours}$ & \textbf{36.83} & \textbf{54.62} & \textbf{30.06} & \textbf{47.61} \\
        Proto$_{AB1}$ & / & / & & 1.04\\
        Proto$_{AB2}$ & / & 9.89 & / & /\\
        \midrule
        \multirow{2}{*}{\textbf{Model}} & \multicolumn{4}{c}{\textbf{FEW-NERD(INTER) F1 scores}}\\
        \cmidrule(lr){2-2}\cmidrule(lr){3-3}\cmidrule(lr){4-4}\cmidrule(lr){5-5}
        & \textbf{5 way 1\textasciitilde2 shot} & \textbf{5 way 5\textasciitilde10 shot} & \textbf{10 way 1\textasciitilde2 shot} & \textbf{10 way 5\textasciitilde10 shot}\\
        \midrule
        Proto$_\mathrm{ours}$ & \textbf{54.35} & \textbf{66.93} & \textbf{47.32} & \textbf{61.50} \\
        Proto$_{AB1}$ & 9.24 & / & / & /\\
        Proto$_{AB2}$ & / & / & 1.56 & /\\
        \bottomrule
    \end{tabular}
    \end{threeparttable}
    \label{tab:few-nerd-ab}
\end{table}

\end{document}